\documentclass[runningheads]{llncs}

\usepackage{eccv}

\usepackage{eccvabbrv}

\usepackage{graphicx}
\usepackage{booktabs}
\usepackage{amsmath}
\usepackage[accsupp]{axessibility}  % Improves PDF readability for those with disabilities.

\newcommand{\db}[1]{{\color{orange}{db: #1}}}

\newcommand{\mycomment}[1]{}
\renewcommand{\inst}[1]{\textsuperscript{#1}}

\usepackage{hyperref}

\usepackage{orcidlink}

\begin{document}

\title{Unlocking Comics: The AI4VA Dataset for Visual Understanding} 

\titlerunning{Unlocking Comics}

\author{Peter Gr\"onquist\inst{1,2}\orcidlink{0000-0002-3290-9361}\and
Deblina Bhattacharjee\inst{1,3}\orcidlink{0000-0002-0534-852X}\and
Bahar Aydemir\inst{1}\orcidlink{0000-0001-5202-5240} \and
Baran Ozaydin\inst{1}\orcidlink{0009-0000-8512-3381} \and
Tong Zhang\inst{1}\orcidlink{0000-0001-5818-4285} \and
Mathieu Salzmann\inst{1}\orcidlink{0000-0002-8347-8637} \and
Sabine S\"usstrunk\inst{1}\orcidlink{0000-0002-0441-6068}}

%\authorrunning{P.~Gr\"onquist et al.}

\institute{
\inst{1}École Polytechnique Fédérale de Lausanne (EPFL), Lausanne, Switzerland\\
\inst{2}Huawei Technologies, Zürich Research Center, Switzerland \\
\inst{3}University of Bath, UK \\
\email{\{firstname.lastname\}@epfl.ch}}
\maketitle

\begin{abstract}
In the evolving landscape of deep learning, there is a pressing need for more comprehensive datasets capable of training 
models across multiple modalities. Concurrently, in digital humanities, there is a growing demand to leverage technology for diverse media adaptation and creation, yet limited by sparse datasets due to copyright and stylistic constraints.
Addressing this gap, our paper presents a novel dataset comprising Franco-Belgian comics from the 1950s annotated for tasks including depth estimation, semantic segmentation, saliency detection, and character identification. It consists of two distinct and consistent styles and incorporates object concepts and labels taken from natural images. By including such diverse information across styles, this dataset not only holds promise for computational creativity but also offers avenues for the digitization of art and storytelling innovation. 
This dataset~\footnote{Work done when PG, DB, BA, and BO were at EPFL and supported in part by the Swiss National Science Foundation via the Sinergia grant CRSII5-180359.} is a crucial component of the AI4VA Workshop Challenges~\url{https://sites.google.com/view/ai4vaeccv2024}, where we specifically explore depth and saliency. Dataset details at \url{https://github.com/IVRL/AI4VA}.
\keywords{Semantic segmentation \and Depth estimation \and Saliency detection \and Comics \and Dataset}
\end{abstract}

\section{Introduction}
As the fields of computer vision (CV), machine learning, and more specifically deep learning continue to advance, there has been a significant expansion in the domains of image analysis, perception, and understanding~\cite{10.1023/B:VISI.0000029664.99615.94,NIPS2012_c399862d,937709}. This research spans from color representation in digital images and networks~\cite{8296790} to the principles of gestalt~\cite{gestalt}, human perception~\cite{CHEN2007193}, human understanding~\cite{BIEDERMAN198529}, and its counterpart in deep learning: internal state representation~\cite{lecun2015deep}. Amidst these advancements, the analysis of `natural images'~\cite {naturalimages,5206848} has been a driving force behind deep learning research. However, 
recent progress in the fields of scene understanding~\cite{radford2021learning,alayrac2022flamingo}, large language models (LLMs)~\cite{NEURIPS2020_1457c0d6,touvron2023llama}, and generative AI~\cite{rombach2022highresolution} have highlighted the necessity for datasets that encapsulate abstract and complex visual narratives, 
especially those intertwining textual and visual information in innovative ways. %foundational models that grasp concepts from diverse modalities, emphasizing the critical role of datasets, such as CAT2000~\cite{cat2000} 
%TODO needs a connection to generic models/foundational models

Despite these advancements, there remains a notable scarcity of datasets that delve into the abstract complexities of visual narratives, particularly those that blend textual and visual information in unique ways. This gap underscores the untapped potential of comics as a rich source of data for deep learning research. Comics, with their long history of conveying stories through a combination of language and imagery, offer a unique opportunity to study abstract or `symbolic thinking'~\cite{10.3389/fpsyg.2018.00115}. This form of storytelling, ranging from cave paintings and hieroglyphs to modern art and social media, showcases the human capacity to communicate complex concepts through visual means. 
Comics, in particular, uniquely merge the descriptive precision of language with the abstraction of concepts through artistic depictions, that are deeply subjective and creative, thus making them an exemplary source for training generic models to understand abstract concepts.
%We believe that such multi-modal data, blending visual and textual elements, are precisely what future foundational and more generic models need to navigate today's complex digital landscapes, capturing the essence of human communication and cognition.

Recognizing the value of comics in the exploration of visual narratives and such abstract concept understanding, we introduce a novel dataset comprising mid-twentieth-century Franco-Belgian comics, which provides a rich and diverse array of scenes. We select Franco-Belgian comics for curating this dataset because of the scarcity of explorations into European comic styles, 
with existing comics datasets such as the Digital Comic Museum (DCM)~\cite{jimaging4070089}, comic2k~\cite{inoue_2018_cvpr}, COMICS~\cite{IyyerComics2016}, ebdtheque~\cite{eBDtheque}, and Manga109~\cite{multimedia_aizawa_2020} predominantly focusing on English, American or Japanese comics whose styles, storytelling, and narration techniques differ from that of their European counterparts. This focus on Franco-Belgian comics aims to enrich the dataset diversity and honor European cultural heritage.

Our dataset, called AI4VA comprising comics imagery from Placid et Muzo - Yves le loup - Bandes Dessinées, encompasses a wide range of comic styles, from colored and black-and-white realistic images to toonified and abstract scenes. The comics are annotated for semantic segmentation, ordinal depth, and visual saliency, providing a comprehensive resource for studying how visual stories are structured and interpreted across different media. By focusing on the integration of text and abstract imagery, the AI4VA dataset aims to facilitate research into semantic segmentation, depth perception, saliency estimation, and the broader understanding of visual narratives within deep learning models. 

In contrast to existing benchmarks like DCM~\cite{jimaging4070089}, which offers annotations for panel, character, and face recognition; comic2k~\cite{inoue_2018_cvpr}, focusing on object recognition via domain transfer without task-specific annotations; COMICS~\cite{IyyerComics2016}, centered on narration; Manga109~\cite{multimedia_aizawa_2020}, aimed at retargeting, colorization, and dialog detection; and ebdtheque~\cite {eBDtheque}, which specializes in speech-balloon recognition, AI4VA introduces a broader range of annotations to support a wider array of computer vision tasks.

AI4VA thus fills a critical gap in the availability of complex, narrative-rich visual datasets for advancing deep learning's capability to interpret and generate visual stories. Through this contribution, we aim to catalyze further research into the intricate dynamics of text-image interplay and the cognitive processes underlying the perception and understanding of visual narratives. In summary, our contributions are as follows:
\vspace{-7pt}
\begin{enumerate}
    \item \textbf{AI4VA Dataset:} We present the AI4VA dataset, a first-of-its-kind collection of mid-20th-century Franco-Belgian comics, showcasing a range from realistic to abstract comic styles, filling a gap in European comic research.
    \item \textbf{Rich Annotations for Computer Vision:} We provide detailed annotations for semantic segmentation, ordinal depth, and visual saliency, enhancing capabilities for extensive visual narrative analysis beyond the scope of existing datasets focused on basic comic elements.
    \item \textbf{Advancing Text-Image Interplay Research:} 
With AI4VA we establish a benchmark which we expect to open up further explorations on text-image dynamics within abstract visual contexts, enhancing deep learning's grasp and creation of visual narratives. %, thus expanding AI's exploration of visual storytelling and cognitive processes.
\end{enumerate}
\vspace{-20pt}
\section{Related Work}
There have been several datasets analyzing comics and semantic understanding, such as comic2k~\cite{inoue_2018_cvpr} which makes use of different domains, including watercolor, comic and clip-art to analyze and train models for object recognition with domain transfer. Another dataset, COMICS~\cite{IyyerComics2016}, takes its data from the ``golden age of American comics'' (1938–1954) using the 4,000 highest rated comics on the Digital Comics Museum (DCM\footnote{\url{https://digitalcomicmuseum.com/}}). One of their aims is to allow a model to train on understanding cues from both visual and textual clues, to infer what scene should logically follow, which allows for a more thorough contextual understanding of the story. 

Focusing on a different style, Manga109~\cite{multimedia_aizawa_2020} was one of the first and larger datasets to be published. Originally introduced for sketch-based retrieval, it has since been constantly updated on different tasks such as retargeting, colorization and dialog detection. It is built on 109 comic books of 21,142 pages in total and generally comes closest to what we aim to achieve through a full semantic understanding of comics via automated methods.

Following this is DCM~\cite{jimaging4070089}, another dataset based on the DCM library. This dataset focuses on 772 annotated images from 27 Golden Age comic books, all from different publishers, and is annotated with bounding boxes for the panels, characters, and faces. Within larger databases of comics, models capable of detecting panels, characters, and faces could then help with indexing and searching for semantic content within images and not only through text. 

These datasets themselves already delve deep into understanding comics, however with the recent advances in semantic understanding and style transfer, they do not contain enough information, for example, relating to depth, character-object interaction, and segmentation. Additionally, the comics themselves focus explicitly on American or Japanese comics, and as of writing this paper, we have yet to find a source for Franco-Belgian comics. On top of that, many of these comics are behind copyright claims, making it even harder to perform open-source research on them and publish them. This is why we, in turn, also focus on older Franco-Belgian comics, which have since then entered the public domain.
\section{Dataset Collection and Preparation}
\subsection{Data Collection}
\begin{figure}[tb]
  \centering
  \begin{subfigure}{0.49\linewidth}
    \centering
    \includegraphics[height=6.5cm]{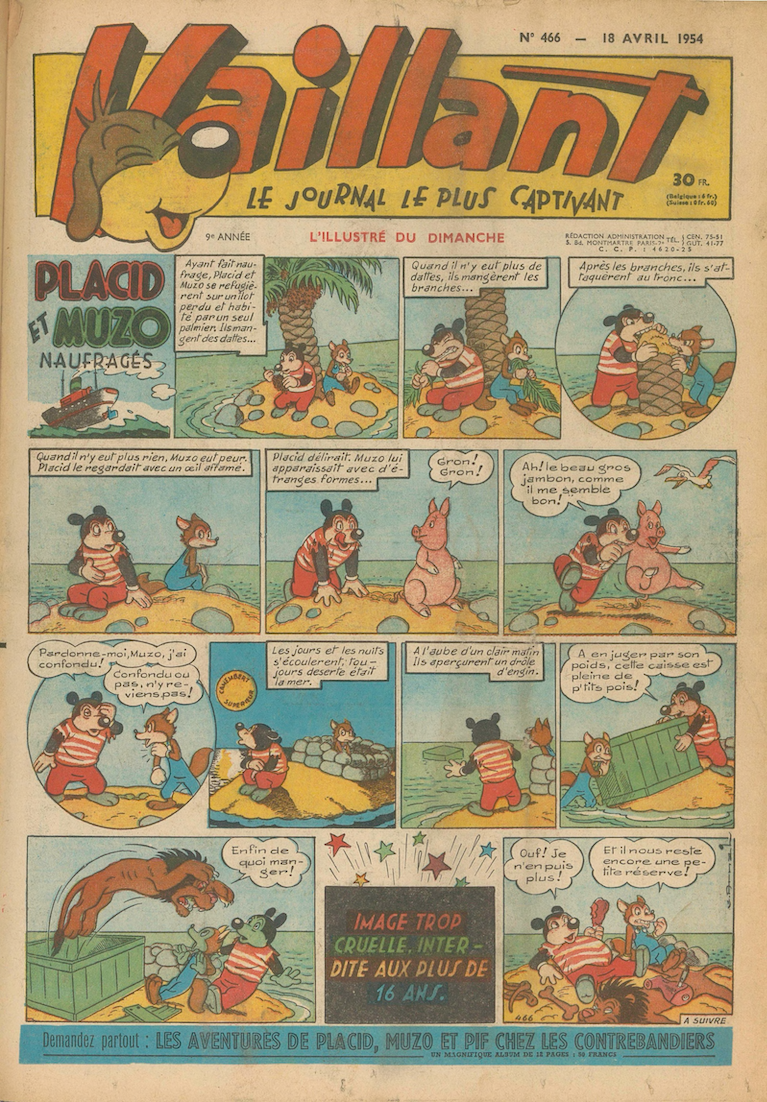}
    \caption{`Placid et Muzo', a whimsical comic featuring abstractly drawn, humanoid animals, a bear and a fox, as the main characters.}
    \label{fig:Placid}
  \end{subfigure}
  \hfill
  \begin{subfigure}{0.49\linewidth}
    \centering
    \includegraphics[height=6.5cm]{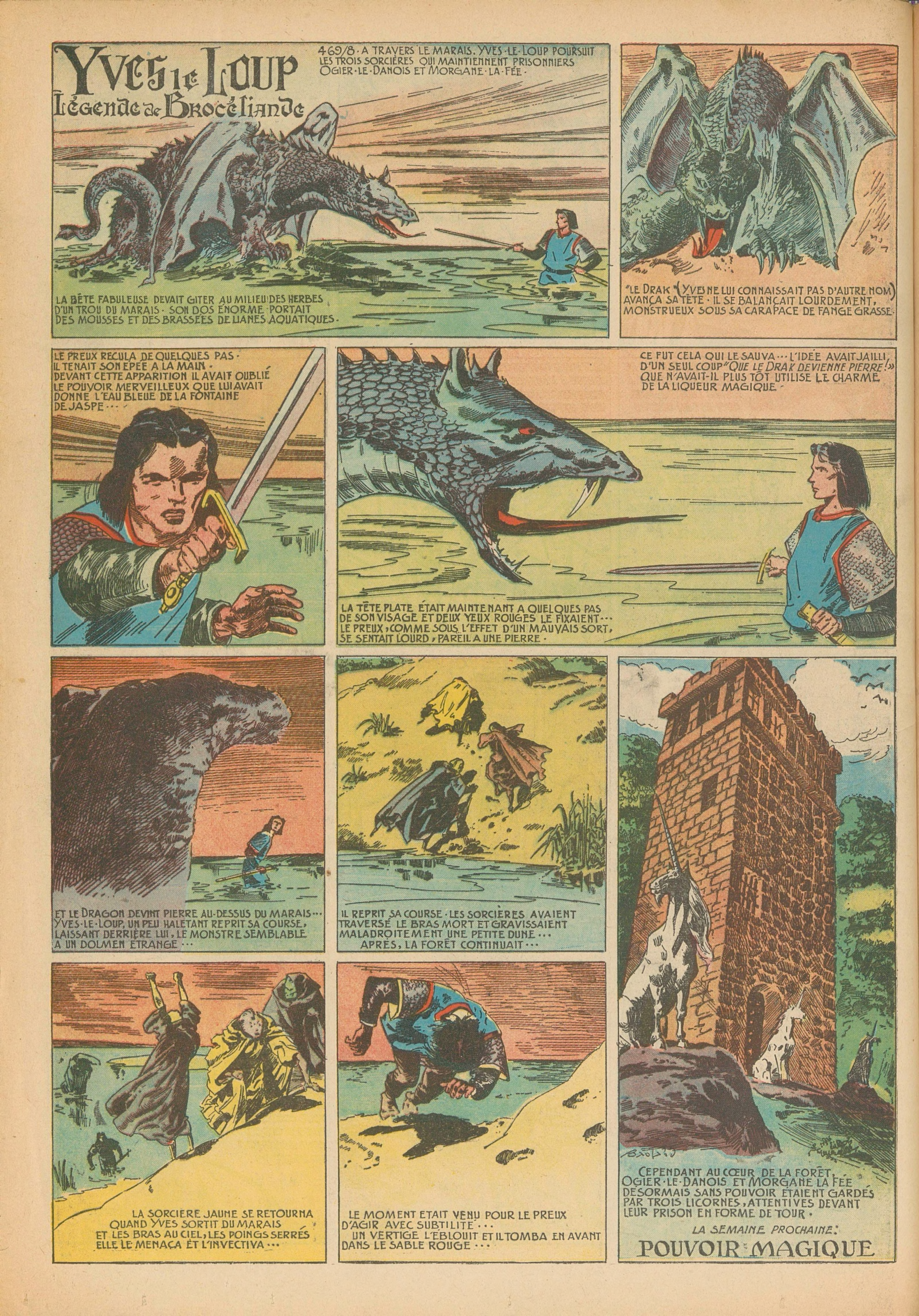}
    \caption{`Yves le loup', a more realistic, intricately drawn historical fantasy mix, centering around Yves, a warrior, as protagonist.}
    \label{fig:Yves}
  \end{subfigure}
  \caption{Example images from both comics in our dataset.}
  \label{fig:PYBD}
\end{figure}

Our dataset encompasses 282 pages from two distinct series featured in `Vaillant', a French weekly journal. `Placid et Muzo', accounting for 154 pages, offers a whimsical narrative of a fox and bear duo, spanning from 1946 to the late 1950s, and is noted for its multiple redrawings and reinterpretations. `Yves le loup', comprising 128 pages, delivers a realistically drawn historical fantasy centered on warrior Yves, serialized from 1947 to 1965 (See Fig~\ref{fig:PYBD}).

Sourced from the Comic Center of Lausanne (Centre BD\footnote{\url{https://www.lausanne.ch/vie-pratique/culture/bibliotheques-et-archives/centre-bd.html}}),  
%one of Europe's most 
we digitized `Vaillant' issues from 1954 to 1957 to curate the AI4VA dataset. This subset encapsulates the full range of dates accessible to us, whilst also highlighting two distinct artistic styles. The dataset features a mix of serialization styles and narratives, including special color editions, traditional black and white, and editions with a unique red tint.

The original image quality is preserved in the digitization, reflecting the historical aging of the pages, such as color fading and yellowing. Each page is provided in PNG format, named after the journal issue, publication date, and page number, ensuring detailed documentation for research use.
%Our dataset consists of 282 pages of comics, gathered from the comics 'Placid et Muzo' a whimsical and abstract series focusing on the tales of the similarly named duo of fox and bear, running from 1946 towards the end of the 1950s, which has been redrawn and reinterpreted a countless number of times, and 'Yves le loup' a realistically drawn historical-fantasy drama, based on the adventures of warrior Yves, which ran from 1947-1965. 
%Both were serialized in the magazine 'Vaillant', a weekly french journal running diverse comic series. Through the large collection of the Comic Center of Lausanne (Centre BD\footnote{\url{https://www.lausanne.ch/vie-pratique/culture/bibliotheques-et-archives/centre-bd.html}}), one of the largest in Europe, we were able to scan their collection of the Vaillant serialization and are now publishing this subset of it, specifically the ranges from 1954-1957, fully annotated. This subset constitutes the full range of dates available to us, as well as two distinct artistic styles.

%Of the 282 pages, 154 are from 'Placid et Muzo' and 128 are from 'Yves le loup', the pages diverge due to the different serialization and stories of each. Additionally, some are special color editions, whilst other pages are in black and white or even other coloring schemes using red tint only.
%The pages themselves are provided in their original scan quality, which includes a certain color deterioration and yellowing of pages.
%Each image is provided in png format, with the journal number, its date and the page number as name.

%TODO can be removed/size-adapted depending on necessity

\subsection{Data Cleaning}
The dataset was carefully curated from the available collection of `Vaillant' comics, specifically including pages that feature `Placid et Muzo' or `Yves le loup' within the chosen time-frame. Notably, the dataset exhibits a degree of noise resulting from the original printing processes, subsequent scanning, and the natural degradation of the print materials. We decided against employing basic image restoration techniques to maintain the material's integrity, thus supporting a wider range of analytical methodologies and to also enable research in comics restoration. The manual annotation process was executed in three stages: Initial segmentation of instances, followed by annotation of saliency and depth, and concluded with a verification phase to ensure uniformity in annotation quality. 
% Please move to GitHub readme: However we do note that there can be slight discrepancies in the annotations, due to the number of annotators involved.

%The data has been manually selected from the original set of 'Vaillant' comics available and contains all pages that include either a 'Placid et Muzo' or 'Yves le loup' panel from within the selected time-period. The data is also relatively noisy, due to printing, scanning and degradation of the printed material, however we have opted not to remove that noise through basic processing means, to preserve the original rendition of the material, allowing for more general approaches to the problem.
%We do find that for simple networks, denoising and normalizing the color histogram can lead to large improvements in segmentation accuracy and classification. 
%The annotations themselves were done in three iterations: First the instances were segmented, then annotated for the respective tasks, finally there was a control pass, to make sure the quality is more or less consistent. However we do note that there can be slight discrepancies in the annotations, due to the number of annotators involved. %TODO this was done by me and students on subsets of the dataset, I for now will not add numbers, but if you think that this 100% needs numbers to back it up, then we can also delete this part, unless someone has ablation studies on the full dataset ready.

\subsection{Annotation}
The images within AI4VA were manually annotated over a span of 1200 hours, with each page requiring approximately 4 hours of work, 
%. To mitigate both inter-annotator and intra-annotator variability, the annotation process was executed in three rounds
by a pool of 13 paid annotators. Subsequently, the annotations underwent verification by five experts in the comic domain to ensure consistency and uniformity in annotation quality. The manual annotation of the comics utilized the Computer Vision Annotation Tool~\cite{boris_sekachev_2020_4009388} (CVAT), developed by the Open Source Computer Vision Library (OpenCV). 
The annotation process itself, was split into two different tasks, starting with the segmentation of graphical components, including characters and objects. These segments were then annotated for semantic segmentation, saliency estimation, and depth ordering. %We separate the task into two different steps, firstly the segmentation aspect, followed by the annotation, which in turn is split into semantic, saliency, and depth parts.

\subsubsection{Segmentation}

\begin{figure}[tb]
  \centering
  \hspace*{\fill}
  \begin{subfigure}[t]{0.483\linewidth}
    \centering
    \includegraphics[width=\linewidth]{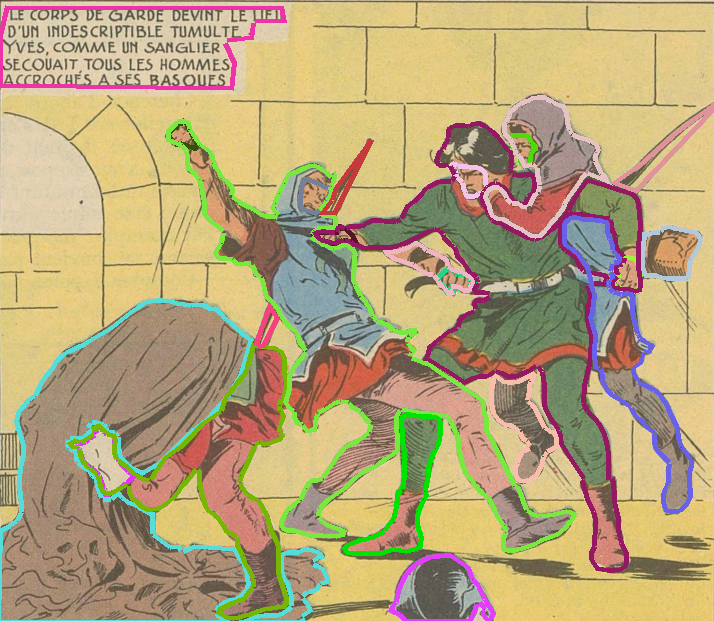}
    \caption{Segmentation as shown in our dataset. Different colors represent different instances. Different instances of the same character are grouped in the annotation (not displayed here).}
    \label{fig:Yseg}
  \end{subfigure}
  \hfill
  \begin{subfigure}[t]{0.49\linewidth}
    \centering
    \includegraphics[width=\linewidth]{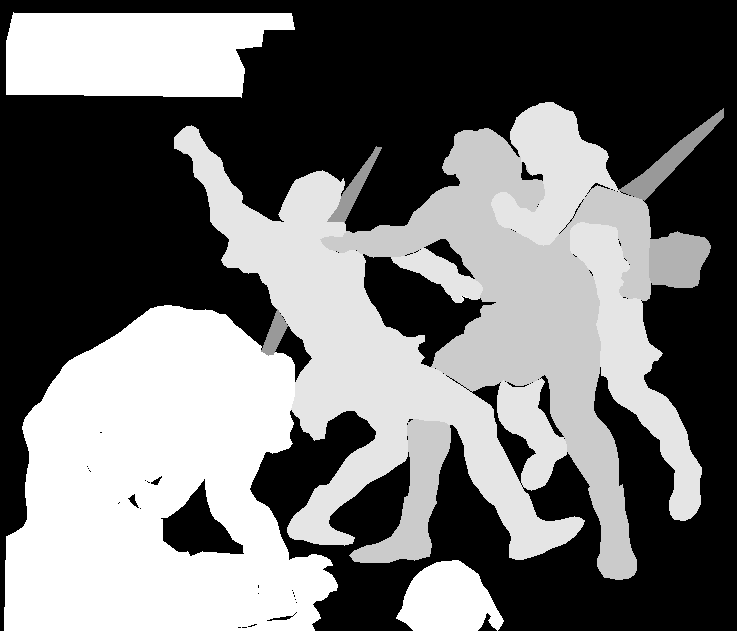}
    \caption{Our ordinal depth annotation, as represented by inter- and intra-depth. Brighter is closer and darker is farther from the viewer. Note that, hands and faces can have different depths from the body of the character.}
    \label{fig:Ydepth}
  \end{subfigure}
  \hspace*{\fill}
  \caption{Annotated panel from `Yves le loup'.}
  \label{fig:segm}
  \vspace{-15pt}
\end{figure}

In our segmentation process, we delineate six distinct categories (see Table~\ref{tb:segmentation annotation}), drawing inspiration from both the Common Objects in COntext (COCO)~\cite{cocodataset} dataset and the unique challenges in visual comprehension we have identified. Our approach includes manually segmenting objects involved in interactions, 
even if they do not fit within predefined categories of COCO, 
%irrespective of their conventional categorization, 
to enhance narrative comprehension of the comic page. Furthermore, we delineate fundamental comic elements such as panels, speech bubbles, and text, with particular attention to the \textit{horizon line}. This line, annotated as a flat polygon, demarcates the visual limit beyond which the depth of objects and characters becomes indistinguishable, as elaborated in our depth annotation methodology. Additionally, we segment faces and hands to facilitate more precise depth annotations (illustrated in Fig.~\ref{fig:Yseg}).
%In regards to the segmentation, we separate the task into segmenting six different branches (See Table~\ref{tab:segmentation}). These elements are partially based on the Common Objects in Context (COCO) dataset~\cite{cocodataset}, and partially on the unique aspects in regards to our challenge.
%We specifically segment objects that are interacted with, even if they do not fit within other categories, as we consider them useful to understanding the general narrative of the page. Additionally, we segment basic comic elements such as panels, speech bubbles and text, with the special note of the horizon line. We consider the horizon line in comics (annotated as a flat polygon), the line beyond which, any difference in depth of objects and characters is not perceivable anymore, as described in the depth annotation part. Finally, we also segment the face and the hands, allowing for a more distinct depth annotation (See Fig.~\ref{fig:Yseg}).

%(TODO Mention inpainting also)

\begin{figure}[tb]
  \centering
  \hspace*{\fill}
  \begin{subfigure}[t]{\linewidth}
    \centering
    \includegraphics[width=\linewidth]{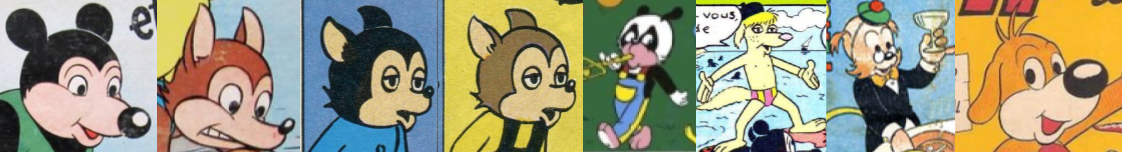}
    \caption{Characters in `\textbf{Placid et Muzo}': \{"Generic":0, "Placid":1, "Muzo":2, "Riri":3, "Fifi":4, "Tib":5, "Mecton":6, "Professeur Grostalent":7, "Pif le Chien":8\}}
    \label{fig:Pchar}
  \end{subfigure}
  \hfill
  \begin{subfigure}[t]{\linewidth}
    \centering
    \includegraphics[width=0.1\linewidth]{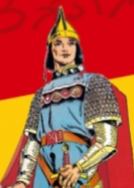}
    \caption{Characters in `\textbf{Yves le loup}': \{"Generic":0, "Yves":1\}, as there are no other recurring characters.}
    \label{fig:Ychar}
  \end{subfigure}
  \hspace*{\fill}
  \caption{Characters of the respective comic serialization with their ID.}
  \label{fig:chars}
  \vspace{-15pt}
\end{figure}

\begin{table}
  \label{tab:segmentation}
  \centering
  \scalebox{0.7}{
  \begin{tabular}{l|l}
    {\bf Category}&{\bf Elements}\\
    \hline \hline
    {\bf Furniture}&Chair, sofa, table\\
    \hline
    {\bf Characters}&Character, face, hand\\
    \hline
    {\bf Animals}&Cat, dog, bird, sheep, cow, horse, generic animal\\
    \hline
    {\bf Vehicles}&Car, bus, bicycle, motorbike, airplane, boat, train\\
    \hline
    {\bf Objects}&Building, plant\\
    \hline
    {\bf Generic objects}&\textit{Other kinds of} objects \textit{that the characters}\\
    &\textit{physically touch, hold or utilize in the page}\\
    \hline
    {\bf Comic structure}&Panel, comic bubble, text, horizon line\\
  \end{tabular}}
  \caption{Semantic categories of segmentation.}
   \label{tb:segmentation annotation}
   \vspace{-15pt}
\end{table}
In the `Placid et Muzo' series, the characters are depicted as humanoid animals, and therefore we classify all of them under the character category rather than animals. In instances where these characters interact with non-humanoid or actual animals, those entities are annotated according to their specific animal classifications. The criterion for an entity to be considered a character is its bipedal stance and ability to speak.
%Characters in 'Placid et Muzo' are all based on animals/humanoid animals, therefore we annotate them all as characters and not animals. There are scenes when characters interact with actual animals, \db{in which case, such animals are annotated as per their animal category. The distinction is that the animal has to be bipedal and speaking, for it to be classified as character.}
Additionally, each recurring character receives a unique identifier- "Character\_ID", which allows for the training models to recognize the characters in different scenes (See Fig.~\ref{fig:chars}). 

%TODO add images for each if space
%For '\textbf{Placid et Muzo}': \{"Generic":0, "Placid":1, "Muzo":2, "Riri":3, "Fifi":4, "Tib":5, "Mecton":6, "Professeur Grostalent":7, "Pif le Chien":8\}

%For '\textbf{Yves le loup}': \{"Generic":0, "Yves":1\}, as there are no other recurring characters.


\subsubsection{Depth}
Ordinal depth involves a qualitative ranking of objects by their distance from the observer, categorizing them as either "closer" or "further away" without quantifying the exact distances. This contrasts with metric depth, which quantifies the exact distances between the observer and objects within a scene, typically in units such as meters or centimeters. Human perception is generally poor at accurately estimating metric depth or the three-dimensional metric structure from a single viewpoint. This is because metric depth in a single image is inherently ambiguous; for example, a tree positioned behind a house might appear larger yet be further away, or smaller and closer, making it impossible to determine the absolute depth difference between the two objects uniquely. Moreover, even when humans can gauge metric depth to some extent, extracting precise numerical values from such estimations remains problematic.

Given these challenges, humans are more adept at assessing relative depth, finding it easier to answer questions like "Is point A closer than point B?". Based on this premise, we opt to manually annotate the AI4VA dataset for ordinal depth, recognizing that accurate metric depth annotations are particularly difficult to obtain in monocular settings, such as with comic images. Specifically, we use ordinal and integer-based \textit{inter-} and \textit{intra-depth} planes. The inter-depth planes, denoted as $[0,\infty)$, are utilized to represent broader distance intervals between distinct ordinal levels. 
To then capture subtler variations within a given inter-depth plane, we utilize intra-depth planes, defined within the range $[1,9]$. This dual-plane system facilitates the delineation of occlusion relationships and the sequential arrangement of each segmented component within a comic panel, including detailed elements like faces and hands (See Fig.~\ref{fig:Ydepth}).
%Coarse planes, which we call inter-depth $[0,\infty)$, show larger distances between the different ordinal planes, whereas to depict more fine-grained differences within an inter-depth plane we use the intra-depth $[1,9]$. Together they give information about occlusion and relative ordering of each segmented element in the panel, including faces and hands (See Fig.~\ref{fig:Ydepth}).
%needs rewrite later

\subsubsection{Saliency}
\begin{figure}[tb]
  \centering
  \hspace*{\fill}
  \begin{subfigure}[t]{0.49\linewidth}
    \centering
    \includegraphics[width=\linewidth]{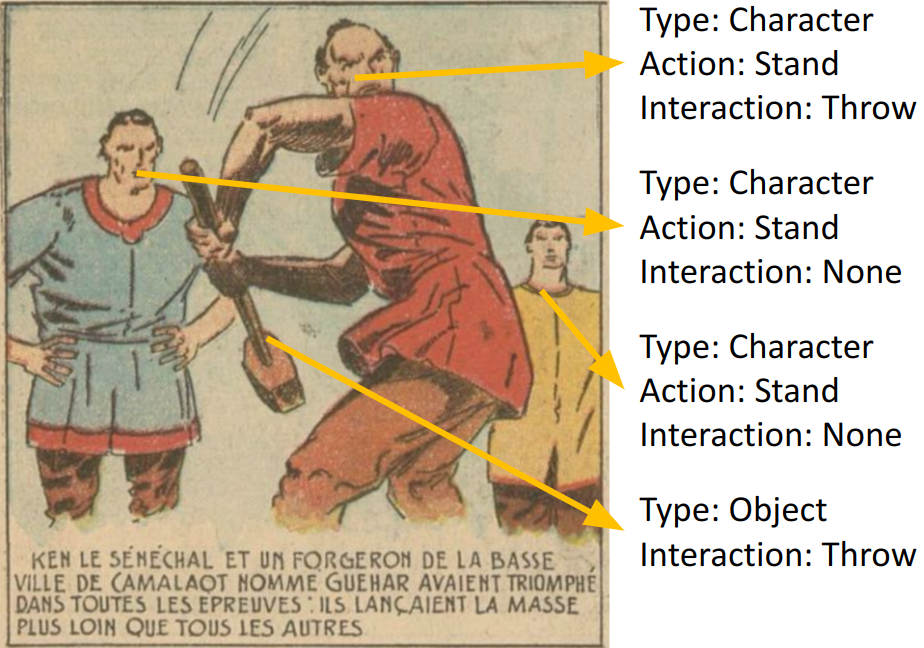}
    %\vspace{-20pt}
    \caption{Example of actions and interactions annotations for a panel of `Yves le loup'. In this case, the hammer plays an important role in the panel and is, therefore, also segmented, labeled as generic object, and annotated as a `Tool'.}
    \label{fig:sal_1}
  \end{subfigure}
  \hfill
  \begin{subfigure}[t]{0.49\linewidth}
    \centering
    \includegraphics[width=\linewidth]{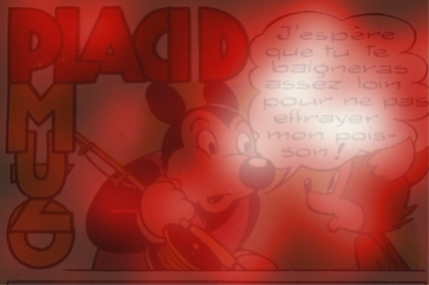}
    \caption{Example of visual saliency from our eye-tracking user study on a panel from `Placid et Muzo'. We cluster the recorded gaze points to form fixations, which we then smooth out by applying a Gaussian filter. White indicates the regions with the highest saliency, transitioning into red and then black for areas without any fixations.}
    \label{fig:sal_2}
  \end{subfigure}
  \hspace*{\fill}
  \caption{Saliency annotations from our dataset}
  \label{fig:saliency_figures}
  \vspace{-15pt}
\end{figure}

In our approach to annotating saliency, we first focus on elements indicative of salient elements, such as various interactions. A fundamental aspect of this is the `tool' attribute, which is marked as true when an object is actively engaged by a character within a scene, and false otherwise.
Furthermore, we provide detailed annotations that capture the specific \textit{action} performed by a character and the \textit{interaction} between a character and an object. \\
\textit{Actions} include: \{"Generic", "Jump", "Lay", "Talk", "Work", "Run", "Smile", "Stand", "Walk"\} \\
\textit{Interactions}: \{"Generic", "Catch", "Cut", "Eat/Drink", "Hold", "Hit/Kick", "Look(at)", "Point", "Read", "Ride", "Skateboard/Ski/Snowboard", "Sit", "Throw", "Carry"\}\\
These annotations aim to identify elements within a panel that significantly contribute to the narrative of the story. For instance, a hammer's saliency is heightened if it is being thrown by a character rather than merely serving as part of the background decor (illustrated in Fig.~\ref{fig:sal_1}).
%\begin{figure}[h!]
  %\centering
  %\includegraphics[width=0.5\linewidth]{Figures/sal_action.png}
  %%\vspace{-20pt}
  %\caption{Example of actions and interactions annotations for a panel of 'Yves le loup'. In this case, the hammer plays an important role in the panel and is, therefore, also segmented, labeled as generic object, and annotated as a 'Tool'.}
%  \label{fig:sal_1}
%\end{figure}
%TODO could use another graphic here, maybe a saliency graphic, and then this can be made into a saliency figure, with both types of saliency displayed

\textbf{Eye tracking user study.} 
Separately, on a subset of our selected comics, we explore the attention patterns of different readers in a saliency user study.
In this eye-tracking study, participants were asked to read and comprehend pages from the AI4VA comic dataset.
Each page was followed by comprehension questions to evaluate their understanding. 
To gather the eye-tracking data, we used a portable eye-tracker~\cite{PupilLabsCore}. This wearable technology enabled participants to move freely during the experiment. Data recording and extraction were conducted using the device's accompanying software. 
We also note that as the comics were written in an older form of French, particularly in the case of `Yves le loup', we ensured beforehand that participants possessed a comprehensive understanding of the language.

\textbf{Creation of ground truth saliency maps.} 
The eye-tracker captures data as gaze points. 
%, which reflect where the participant is looking while reading the comic pages.
%From the eye-tracking devices we receive each participants' gaze points. 
%Initially, we record gaze points through eye-tracking devices as participants interact with a comic page. 
These gaze points pinpoint the exact locations on the visual stimulus that capture viewers' attention at various moments throughout their reading experience. We base our analysis on this collection of gaze points.
By aggregating the raw gaze points data into fixations and applying Gaussian blurring, we transform them into saliency maps, following the methodology described in~\cite{blurringsaliency}. These saliency maps are then created for each page, visually depicting the regions on a comic page that attract the most attention as we illustrate for one panel in Fig.~\ref{fig:sal_2}).
%We then create saliency maps for each page, visually depicting the regions on a comic page that attract the most attention. To transform the raw gaze points data into saliency maps, we aggregate the recorded gazepoints into fixations and apply gaussian blurring following the methodology described in~\cite{blurringsaliency} (See Fig.~\ref{fig:sal_2}).
 %We first collect gaze points recorded by our eye-tracking equipment during participants' engagement with a comic page. These gaze points represent specific locations on the visual stimulus where individuals' eyes are fixated or directed at distinct moments during the reading process. The cumulative dataset of gaze points forms the basis for our subsequent analyses. We generate page-wise saliency maps which are a visual representation of the areas within a comic page that attract the most attention. We convert raw gaze point data into saliency maps by blurring the accumulated gaze points with as in~\cite{blurringsaliency}.\\

\textbf{Panel-wise saliency annotation.}
We recognize that due to the inherent structure of comics, individual panels represent distinct narrative elements and are often consumed by themselves. Therefore, aside from page-wise data, we also present the saliency data in panel-wise form, 
allowing an analysis of how readers allocate attention across both the macroscopic and microscopic levels of a comic narrative.
%Recognizing the inherent structure of comics with distinct panels, we refine our analysis by aligning the saliency maps with panel dimensions. We crop the page-wise saliency maps based on the individual dimensions of each panel. Therefore, we create panel-wise saliency annotations that provide granular insights into the attentional dynamics at the level of each panel. The resulting page-wise and panel-wise saliency annotations offer a valuable framework for understanding how readers allocate attention across both the macroscopic and microscopic levels of a comic narrative. %Therefore, we obtain a deeper comprehension of the cognitive processes underlying comic reading experiences.

For all additional information regarding our eye-tracking methodology, we defer the readers to the supplementary material.

\section{Dataset Statistics and Analysis}
\begin{figure}[tb]
  \centering
  %  \scalebox{0.5}{
  \hspace*{\fill}
  \begin{subfigure}[t]{\linewidth}
    \centering
    \includegraphics[width=\linewidth]{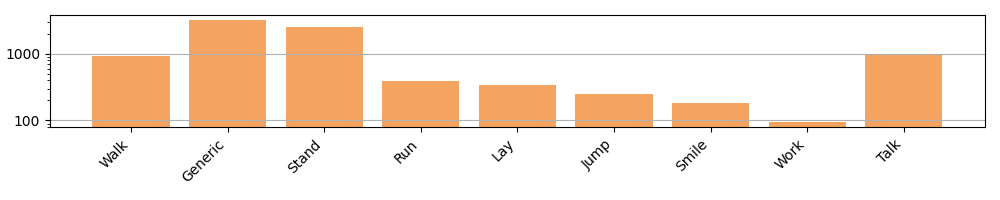}
    \caption{Distribution of character actions.}
    \label{fig:act_hist}
  \end{subfigure}
  \hfill
  \begin{subfigure}[t]{\linewidth}
    \centering
    \includegraphics[width=\linewidth]{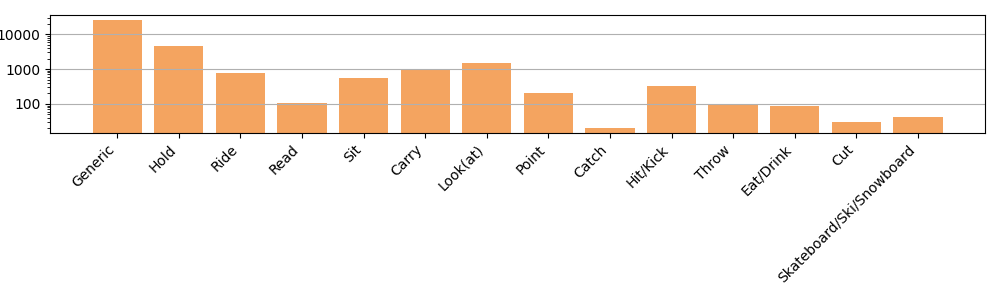}
    \vspace{-40pt}
    \caption{Distribution of object and character interactions.}
    \label{fig:int_hist}
  \end{subfigure}
  \hspace*{\fill}%}
  \vspace{-20pt}
  \caption{Histograms of saliency annotations `action' and `interaction' across AI4VA.}
  \label{fig:sal_hist}
\end{figure}

%General description
%\textbf{Descriptive Statistics}
The AI4VA dataset encompasses a total of 3767 panels, revealing a skewed distribution of classes, particularly noticeable in `Yves le loup' which is a medieval fantasy series, in contrast to `Placid et Muzo' which is set in a contemporary environment featuring diverse landscapes (as shown in Fig.~\ref{fig:act_hist}). This discrepancy is evident in the absence of modern elements like cars or motorbikes in `Yves le loup'. Additionally we find certain COCO classes completely missing, with the full dataset containing only a single instance of a train and no occurrences of sheep.

Analysis of salient actions within the dataset indicates that standing, talking, and walking are the predominant activities among characters, with `generic' actions— encompassing various activities not specifically categorized— being the most frequent. This generic category serves as a placeholder for actions that do not fall into predefined categories, such as `Sitting' or `Dancing' (shown in Fig.~\ref{fig:act_hist}). In terms of object and character interactions, the `generic' term is the most utilized (see Fig.~\ref{fig:int_hist}), accounting for 25050 instances. However, this category can generally be ignored as it represents the default classification for any non-specific interaction, even in scenarios with a lack of interaction. 
Meaning that unless an object or character is annotated as `tool', a `generic' interaction can be considered as `no interaction'. 
Notably, there are 4092 instances labeled as `tool', indicating direct contact between characters and objects, out of a total of 34278 objects/characters identified.
%For example, a generic interaction can be taken to mean 'no interaction' unless the object or character is annotated as a`tool'. 

\begin{figure}[tb]
  \centering
  \scalebox{0.7}{
  \includegraphics[width=\linewidth]{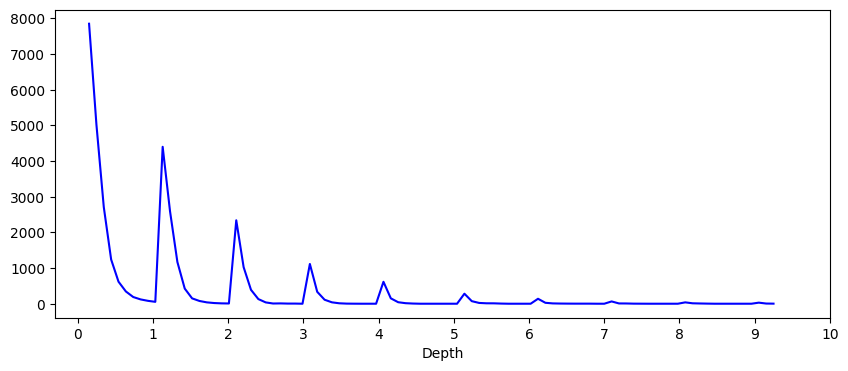}}
  \caption{Distribution of instances by depth: The integers indicate inter-depth ranges from $[0,\infty)$, while the values following the comma specify intra-depth within each inter-depth, ranging from $[1,9]$.}
  \label{fig:depth_hist}
  \vspace{-15pt}
\end{figure}

The dataset employs an ordinal depth scale, predominantly positioning objects within the zero planes of both inter-depth and intra-depth. Each panel invariably includes at least one object in these planes, serving as a reference point for determining the relative depth of subsequent objects. This arrangement goes on iteratively for each depth-plane beyond the first, resulting in a geometric distribution pattern for depth across the dataset, with distinct spikes, representing the inter-depth progression, observable in the depth distribution histograms as shown in Fig.~\ref{fig:depth_hist}.

\begin{figure}[tb]
  \centering
  \hspace*{\fill}
  \begin{subfigure}[t]{\linewidth}
    \centering
    \includegraphics[width=\linewidth]{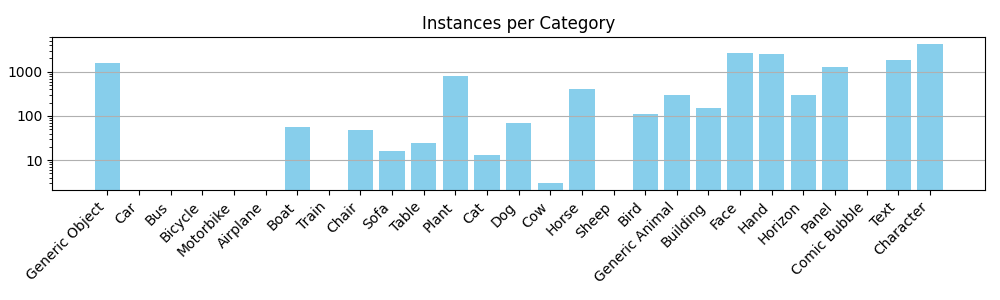}
    \caption{`Yves le loup'.}
    \label{fig:cat_y}
  \end{subfigure}
  \hfill
  \begin{subfigure}[t]{\linewidth}
    \centering
    \includegraphics[width=\linewidth]{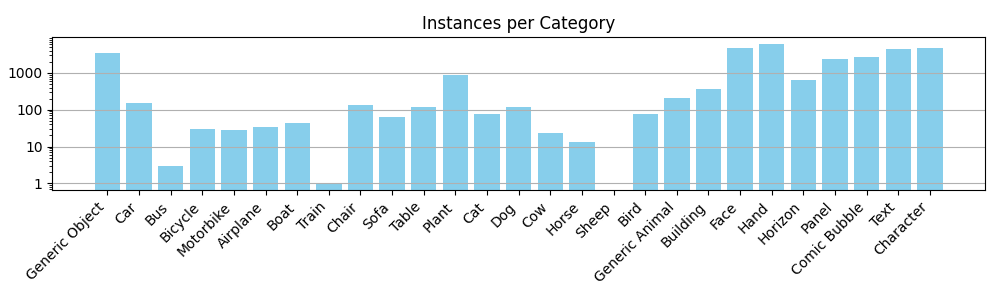}
    \caption{`Placid et Muzo'.}
    \label{fig:cat_p}
  \end{subfigure}
  \hspace*{\fill}
  \caption{Number of instances per category in AI4VA, logarithmic scale.}
  \label{fig:cat}
\end{figure}

%Overall distribution, can be added back to replace separate distribution if not enough space
%\begin{figure}[tb]
%  \centering
%  \includegraphics[width=\linewidth]{Figures/categories_log.png}
%  \vspace{-20pt}
%  \caption{Number of instances per category in AI4VA, logarithmic scale.}
%  \label{fig:cat}
%\end{figure}

%\subsection{Comparative Analysis}

\section{Experiments and Results}
We test the performance of popular vision models (see Sec~\ref{sec:baselines}) on comics scenes from the AI4VA dataset for the tasks of  depth estimation and saliency estimation. This is followed by a comprehensive discussion of the results in Sec~\ref{sec: quantitative results} and Sec~\ref{sec: qualitative results}.
\subsection{Experimental Setup}
\label{sec:experimental setup}
The AI4VA dataset enables benchmarking vision models for two popular vision tasks:  Depth estimation, and saliency estimation. We study the performance of popular methods for each of these tasks by evaluating them against the manual annotations of AI4VA. We use the standard evaluation metrics of Spearman's rank correlation coefficient ($\rho$), Kendall's Tau coefficient ($\tau$), and weighted Kappa ($\kappa$) for depth estimation; and Kullback-Leibler Divergence (KLD), Pearson's correlation coefficient (CC), as well as the Similarity (SIM) score for saliency estimation. 

For the task of \textbf{depth estimation}, we train the existing models (c.f. Sec~\ref{sec:baselines}) under two settings where: 1) the training data comprises the official training split of COCO as well as 2500 AI4VA comics images that are translated to the real-world domain using an unsupervised image-to-image translator named DUNIT~\cite{dunit}; and 2) the training data consists of COCO training images, exclusively. For the first setting, all the depth estimators are trained on the COCO training images utilizing pseudo ground-truth depth data derived from MIDAS~\cite{midas} as well as the AI4VA images using manual ground-truth annotations. Despite the depth models being designed to predict metric depth, our objective is to evaluate their performance in estimating ordinal depth. To achieve this, we adopt the methodology outlined by Zoran et al.~\cite{zoran}, which involves sampling point pairs from the training images (comprising images with semantic labels) through superpixel segmentation. The ordinal relationships between these point pairs are established by comparing their ground-truth depths as determined by MIDAS and the AI4VA manual annotations. The same procedure is applied to the test set to generate the point pairs for evaluation (around 3K pairs per image). 
Note that during inference, we evaluate the ordinal depth predictions on \textit{translated} test comic images from AI4VA.

To evaluate the performance of existing \textbf{saliency estimation} models on our dataset (c.f. Sec~\ref{sec:baselines}), we first translate AI4VA images to the real-world photographic domain using DUNIT~\cite{dunit}, and then assess the prediction of the saliency models on the translated AI4VA images. 
We do not fine-tune the saliency estimators on AI4VA since the evaluated estimators are not optimized to tackle domain discrepancies and therefore, suffer from performance degradation. %Specifically, the disparate domains of the trained data comprising real-world imagery and the fine-tuned data consisting of comic imagery degrade the performance of the saliency estimators. 
%Detailed training settings for vision models, choice of baselines, and data splits can be found in our supplementary.
Additional details on training and choice of baselines can be found in the supplementary material.

\subsection{Baselines}
\label{sec:baselines}
%\subsubsection{Semantic Segmentation} We use the state-of-the-art baselines DeepLabV3~\cite{chen2017deeblabv3} (CNN~\cite{he2016resnet}, in the supplementary) and Mask2former~\cite{cheng2022mask2former} (ViT~\cite{dosovitskiy2020vit}), implemented in Detectron2~\cite{wu2019detectron2}, and report their performances on our comics dataset. 
\subsubsection{Depth Estimation} We evaluate the performance of state-of-the-art depth estimators including convolutional neural network (CNN)-based methods like MIDAS~\cite{midas}, CDE~\cite{CDE}, and ComicDepth~\cite{ComicDepth}; vision transformer-based methods like DPT~\cite{DPT}, DepthFormer~\cite{DepthFormer}, and ZoeDepth~\cite{ZoeDepth}; as well as diffusion-based methods like DDP~\cite{DDP}, DepthAnything~\cite{DepthAnything}, Marigold~\cite{marigold}, and EVP~\cite{EVP} under the two experimental settings mentioned in Sec~\ref{sec:experimental setup}. Note that the diffusion-based methods leverage additional training data, as outlined in their respective works~\cite{DDP, DepthAnything, marigold, EVP}.
\subsubsection{Saliency Estimation}
We evaluate state-of-the-art saliency estimators including DeepGaze I~\cite{kummerer2015deep}, DeepGaze IIE~\cite{linardos2021deepgaze}, DeepGaze III~\cite{deepgaze3}, SimpleNet~\cite{simplenet}, UNISAL~\cite{unisal}, UMSI~\cite{Nature_computationalModelingOfVisualAttention}, TempSAL~\cite{aydemir2023tempsal} and gradient-based CAM methods~\cite{layercam,zhou2015cnnlocalizationcam,jacobgilpytorchcam,selvaraju2017grad,Omeiza2019SmoothGA} on our saliency ground-truth annotations. \\ \\
%We assess each model’s performance using three primary metrics: Kullback–Leibler divergence, Pearson’s Correlation Coefficient, and Similarity. These metrics facilitated performance comparisons between saliency models and ground truth saliency maps, as well as among different models. 
We provide additional details about all the baselines in the supplementary material.
\subsection{Quantitative Results} 
\label{sec: quantitative results}

\textbf{Depth Estimation.} 
\begin{table}[ht]
\scalebox{0.7}{
\centering
\setlength{\tabcolsep}{5pt}
\begin{tabular}{llcccccc}
& &\multicolumn{3}{c}{\textbf{Setting 1: COCO w/ AI4VA}} & \multicolumn{3}{c}{\textbf{Setting 2: COCO only}}\\
\textbf{Network architecture}&\textbf{Model} & \textbf{$\rho$} $\uparrow$ & \textbf{$\tau$}$\uparrow$  & \textbf{$\kappa$} $\uparrow$ & \textbf{$\rho$} $\uparrow$ & \textbf{$\tau$}$\uparrow$  & \textbf{$\kappa$} $\uparrow$  \\ 
 \cmidrule(l){1-1}\cmidrule(l){2-2}\cmidrule(l){3-5}\cmidrule(l){6-8}
&MIDAS~\cite{midas} & 0.89 & 0.88 & 0.89& 0.70 & 0.70 & 0.69 \\
CNN-based& CDE~\cite{CDE} & 0.88 & 0.89 & 0.87& 0.68 & 0.67 & 0.69\\
&ComicDepth~\cite{ComicDepth} & 0.90 & 0.91 & 0.89 & 0.71 & 0.71 & 0.71\\
 \cmidrule(l){1-1}\cmidrule(l){2-2}\cmidrule(l){3-5}\cmidrule(l){6-8}
&DPT~\cite{DPT} & 0.89 & 0.90 & 0.88& 0.70 & 0.71 & 0.70\\
Vision transformer-based&DepthFormer~\cite{DepthFormer}  & 0.90 & 0.91 & 0.90 & 0.74 & 0.75 & 0.73\\
&ZoeDepth~\cite{ZoeDepth} & 0.91 & 0.92 & 0.90& 0.75 & 0.75 & 0.76\\
 \cmidrule(l){1-1}\cmidrule(l){2-2}\cmidrule(l){3-5}\cmidrule(l){6-8}
&DDP~\cite{DDP} & 0.90 & 0.92 & 0.90& 0.75 & 0.76 & 0.73\\
Diffusion-based&DepthAnything~\cite{DepthAnything}  & \textbf{0.95} &\textbf{ 0.97} & \textbf{0.94}& \textbf{0.79} & \textbf{0.80} & \textbf{0.79}\\
(Extra image-text training data)& Marigold~\cite{marigold}   & \underline{0.93} & \underline{0.94} & \underline{0.91} & \underline{0.77} & \underline{0.77} & \underline{0.78}\\
&EVP~\cite{EVP}   & 0.91 & 0.93 & \underline{0.91}  & \underline{0.77} & 0.76 & 0.77\\
%\hline
\end{tabular}}
\caption{\small{
%Quantitative results of state-of-the-art depth estimators on the AI4VA comics test images w.r.t $\rho$, $\tau$, and $\kappa$, respectively. We show that all the models perform relatively well under both the experimental settings as detailed in Sec~\ref{sec:experimental setup}, therefore, agreeing with AI4VA's ordinal depth annotations. The best results are in bold and the second-best are underlined.
We present the performance of state-of-the-art depth estimators on AI4VA comics test images, evaluated by $\rho$, $\tau$, and $\kappa$, respectively. Under both the experimental settings, as detailed in Sec~\ref{sec:experimental setup}, the models consistently perform well, aligning with AI4VA's ordinal depth annotations. The best results are marked in bold, with the second-best underscored.}}
\vspace{-15pt}
\label{tb:depth quantitative result}
\end{table}
For the results in Table~\ref{tb:depth quantitative result}, we evaluate 1267 AI4VA comics test images for both the Settings 1 and 2. We show that all the models perform well on the AI4VA test images, therefore agreeing with AI4VA's ordinal depth annotations. 
In particular, diffusion models show the highest performance due to the additional training data comprising image-text cues. Their performance is followed by the vision transformer-based models that attend to longer depth orderings. However, as the AI4VA scenes do not comprise a large number of inter-depth planes, the performance of vision-transformer depth estimators is hindered. Conversely, the CNN-based models suffer a bias towards the intra-depth orderings due to their local receptive field and are, therefore, unable to attend to farther-depth planes. An ideal situation would be a model that combines both the powers of CNN and transformers, thus predicting more accurate inter-and intra-depth orders. Nonetheless, all the models show a positive correlation with AI4VA's depth annotations across all three metrics, highlighting the quality of the ground-truth. 
\\
\textbf{Saliency Estimation.}
Table~\ref{tab:saliency_baselines} displays the results for the saliency prediction models.
DeepGazeIIE~\cite{linardos2021deepgaze}, which was trained with multiple image classification backbones for real-world images, outperforms the other baselines across all evaluation metrics. However, all the baseline models significantly underperform in the domain of comics compared to their effectiveness on natural images. This discrepancy highlights the complexity of saliency prediction within the comic domain and underscores the necessity for models that can effectively adapt to the unique characteristics of such visual narratives.

Please note that all segmentation-related experiments are included in the supplementary materials, as the AI4VA challenge primarily focuses on the depth and saliency aspects of the AI4VA dataset. Given the specific focus on depth and saliency, we also provide preliminary baseline explorations for semantic segmentation using CNN and transformer-based models, based on the AI4VA segmentation annotations, in the supplementary.

\begin{table}[ht]
\centering
\scalebox{0.7}{
\setlength{\tabcolsep}{5pt}
\begin{tabular}{lcccccc}
 & \multicolumn{3}{c}{\textbf{Page}} & \multicolumn{3}{c}{\textbf{Panel}} \\ 
\textbf{Model} & \textbf{KLD}$\downarrow$  & \textbf{CC} $\uparrow$ & \textbf{SIM}$\uparrow$  & \textbf{KLD}$\downarrow$ & \textbf{CC} $\uparrow$ & \textbf{SIM}$\uparrow$  \\ 
 \cmidrule(l){1-1}\cmidrule(l){2-4}\cmidrule(lr){5-7}
DeepGaze I & 0.290 & 0.439 & 0.714 & 0.228 & 0.315 & 0.742 \\
DeepGaze IIE & 0.207 & 0.643 & 0.771 & 0.194 & 0.615 & 0.793 \\
DeepGaze III & 0.254 & 0.530 & 0.742 & 0.222 & 0.482 & 0.761 \\
SimpleNet & 0.351 & 0.548 & 0.729 & 0.783 & 0.514 & 0.792 \\
UNISAL  & 0.598 & 0.517 & 0.730 & 0.389 & 0.528 & 0.673 \\
UMSI & 0.495 & 0.463 & 0.629 & 1.994 & 0.428 & 0.580 \\
TempSAL & 0.363 & 0.573 & 0.714 & 1.329 & 0.507 & 0.555 \\
\cmidrule(l){1-1}\cmidrule(l){2-4}\cmidrule(lr){5-7}

CAM  & 0.373 & 0.359 & 0.696 & 0.386 & 0.312 & 0.743 \\
GradCAM   & 0.364 & 0.351 & 0.697 & 0.375 & 0.454 & 0.751 \\
LayerCAM   & 0.350 & 0.370 & 0.703 & 0.338 & 0.452 & 0.755 \\
SmoothGradCAM++   & 0.333 & 0.398 & 0.713 & 0.330 & 0.464 & 0.760 \\
%\hline
\end{tabular}}
\caption{Quantitative results of the state-of-the-art saliency prediction models on page and panel-based metrics.}
\label{tab:saliency_baselines}
\vspace{-20pt}
\end{table}
\vspace{-20pt}
\subsection{Qualitative Results} 
\label{sec: qualitative results}
\begin{figure}[tb]
  \centering
  \scalebox{0.6}{
  \includegraphics[width=\linewidth]{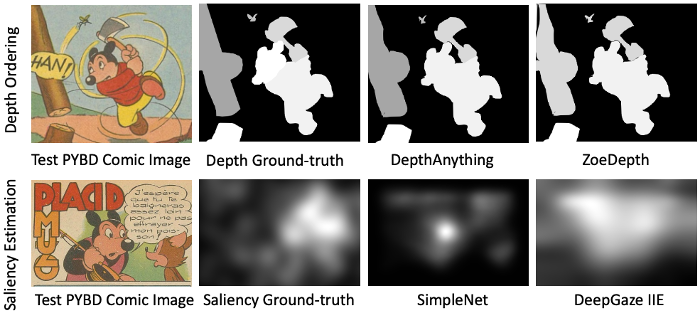}}
  \caption{Qualitative results for depth ordering (top row) and saliency estimation (bottom row) on AI4VA test images from `Placid et Muzo'.  }
  \label{fig:qualitative result}
  \vspace{-15pt}
\end{figure}

 In Fig~\ref{fig:qualitative result}, we show that for depth estimation, the models predict the orders in alignment with that of the ground-truth depth annotations, wherein DepthAnything~\cite{DepthAnything} outperforms the ZoeDepth~\cite{ZoeDepth} method. This corroborates with our findings in Table~\ref{tb:depth quantitative result}. As DepthAnything is a diffusion-based method it leverages extra text-image cues to predict better ordinal depth as compared to the transformer-based ZoeDepth. Further, for saliency, the DeepGazeIIE model ~\cite{deepgaze2} outperforms the SimpleNet~\cite{simplenet} method, due to its in- and out-domain training capabilities, albeit on real-world images. 
\section{Usage and Accessibility}

\subsection{Download and Access}
Instructions for downloading and accessing the dataset are at~\url{https://github.com/IVRL/AI4VA}. 
%The dataset will be released in parts: The first training part will be released by March 20th, 2024. %at the beginning of our challenge in the workshop `AI for Visual Arts Workshop and Challenges' (AI4VA), with parts left out to be used as test-set. After the challenge is over, the full dataset will be published.
We publish our annotations under the Creative Commons (CC) BY license\footnote{\url{https://creativecommons.org/licenses/by/4.0/legalcode}} with Non-Commercial usages.
\vspace{-10pt}
\subsection{Legal Considerations}
The authors and contributors of both comics have all since passed. We have to the best of our ability reached out to all possibly concerned parties and nobody has claimed the rights, therefore the underlying images can be used for unrestricted use in personal research, non-commercial, and not-for-profit endeavours. For any other usage scenarios, kindly contact us via email, providing a detailed description of your requirements.. If you are the owner of the rights of part or all of the comics referenced in this paper and find issue with this, please contact us.

\section{Limitations and Future Work}
In this work, we have introduced a novel dataset featuring abstract Franco-Belgian comic images, comprising annotations that capture ordinal depth, saliency-based elements including actions and interactions, character details, and segmentation. 
%This dataset marks a significant step towards understanding complex visual narratives. 
Despite facing the inherent subjective challenges of comic interpretation—most notably in achieving consistent annotations across varied artistic styles such as for depth perception—we find that the subjective nature of the AI4VA dataset offers unique insights into human cognitive functions. We hope that these insights, particularly regarding perception and understanding, will contribute to advancing our comprehension of how visual narratives are processed.

Future efforts will aim to diversify the dataset with more comic styles and narratives, and refine the annotation process to address the challenges of subjective interpretation. Additionally, we want to develop advanced models for automated comic content generation, as well as enhance existing tools for content format adaptation and accessibility. We thereby hope to foster new creative possibilities, contributing to the fields of computer vision, deep learning, and creative arts. 
To stay updated on this research and any changes to the dataset, please refer to our GitHub repository at~\url{https://github.com/IVRL/AI4VA}.

\textbf{Acknowledgement.} This work was supported in part by the Swiss National Science Foundation via the Sinergia grant CRSII5-180359.

\clearpage  % TODO REVIEW/FINAL: This \clearpage needs to be removed from both review and camera-ready versions.

%\bibliographystyle{splncs04}
%\bibliography{egbib}

\end{document}

\title{Unlocking Comics: The AI4VA Dataset for Visual Understanding (Supplementary)} 

\titlerunning{Unlocking Comics}

\author{Peter Gr\"onquist\inst{1,2}\orcidlink{0000-0002-3290-9361} \and
Deblina Bhattacharjee\inst{1,3}\orcidlink{0000-0002-0534-852X} \and
Bahar Aydemir\inst{1}\orcidlink{0000-0001-5202-5240} \and
Baran Ozaydin\inst{1}\orcidlink{0009-0000-8512-3381} \and
Tong Zhang\inst{1}\orcidlink{0000-0001-5818-4285} \and
Mathieu Salzmann\inst{1}\orcidlink{0000-0002-8347-8637} \and
Sabine S\"usstrunk\inst{1}\orcidlink{0000-0002-0441-6068}}

\authorrunning{P.~Gr\"onquist et al.}

\institute{
\inst{1}École Polytechnique Fédérale de Lausanne (EPFL), Lausanne, Switzerland\\
\inst{2}Huawei Technologies, Zürich Research Center, Switzerland \\
\inst{3}University of Bath, UK \\
\email{\{firstname,lastname\}@epfl.ch}}

\maketitle

The AI4VA dataset is a vital part of the challenges organized for the AI4VA workshop, focusing on two main tasks: depth ordering and saliency estimation. In the following sections, we discuss experiments related to depth ordering, saliency estimation, and semantic segmentation (the latter of which is not included in the AI4VA challenges).

\section{Additional Experimental Details}

\subsection{Depth Estimation}
For the training of our depth estimator across both experimental setups, we initially resize training images to an ``upper-bound'' dimension of $384 \times N$ (or $N \times 384$), where $N$ is a multiple of 32 and falls within the range $32 \leq N \leq 384$, maintaining the original image's aspect ratio. Due to batch processing requirements, all images are standardized to $384\times384$ pixels. This is achieved by ``lower-bound'' resizing images to dimensions where $N \geq 384$, followed by random cropping to obtain a $384\times384$ resolution. The COCO dataset images are preprocessed by translating them into the comic domain using DUNIT~\cite{dunit} to train our models. Training commences with the default hyper-parameters from the original studies, utilizing the Adam optimizer for 100 epochs. Model weights are initialized based on their performance on natural images to preserve pre-learned features, setting a low learning rate of $10^{-6}$. For the experimental setting 1, we employ the L1-loss, comparing the combined pseudo-ground-truth depth for COCO images and the manually annotated ground-truth for AI4VA images against the model's depth predictions. In contrast, setting 2 involves training with an L1-loss between the pseudo-ground-truth depth for COCO images and the predicted depth.

\subsubsection{Translation approach as a baseline  vs. fine-tuning on comics data}
\begin{figure}[h!]
  \centering  \includegraphics[width=1.0\linewidth]{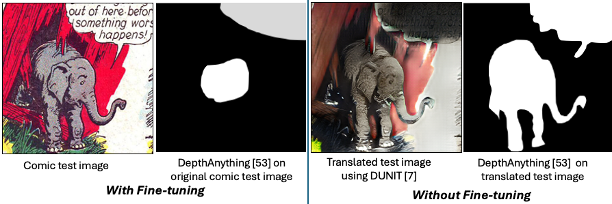}
  \caption{\textbf{[Left]} Depth inference by fine-tuned model on original comics image and \textbf{[Right]} by not fine-tuned model on comics$\rightarrow$real translated image.}
  \label{fig:rebuttal1}
  \vspace{-14pt}
\end{figure}

We extensively experimented with fine-tuning pretrained models directly on comic images, but this approach failed to produce meaningful results for depth and saliency due to the domain disparity between comic data and real-world data. The pretrained models could not adapt to the comic data distribution on top of their learned priors from real-world data. However, when comics data is transferred to the real-world domain, it becomes compatible for inference with models trained on real-world data. To address this domain disparity, we employ DUNIT~\cite{dunit}, which bridges the domain gap through adversarial training, similar to the approach used by ComicsDepth~\cite{ComicDepth}, the first work on depth in the comics domain. In Fig.~\ref{fig:rebuttal1}, we show results for the depth instance on comic images using direct fine-tuning and on translated images. The baselines on the translated images outperform those on the fine-tuned ones. 

\subsection{Saliency Estimation}
\subsubsection{Eye Tracking Details}
At the beginning of our experiments, participants were given instructions about the eye-tracking procedure and had to give their written consent, allowing us to collect and use their data. Furthermore, since the comics are written in an older form of French, particularly in the case of `Yves le loup', we ensured that participants had a comprehensive understanding of the language. Throughout the experiments, participants were required to read and comprehend thirty comic pages of AI4VA. After each page, they were asked questions to assess their understanding. We calibrated the eye tracker every seven images to ensure the quality of the recorded data. One break per experiment session was allowed for the readers.
We displayed the comic pages on a vertical 27-inch screen to match the dimensions of the pages, which are 27x34cm. This arrangement ensured that the images fit perfectly within the screen, providing a more realistic, closer to how the actual comic would be read, viewing experience for the participants. The eye-tracking device~\cite{PupilLabsCore} itself features two infrared eye cameras with a 200Hz sampling rate and a resolution of 192x192 pixels. Furthermore, the device includes a 30Hz 1080p world camera that captures the view of the individual wearing the device. For our experiments, we note that the eye tracker has an angular accuracy of 0.6 degrees which corresponds to 0.98 cm on our screen. We used the provided software to record and extract data.
\vspace{-10pt}
\subsubsection{Gaze points and fixations}
In our experiments, we employ a portable eye-tracking device~\cite{PupilLabsCore} to capture the gaze points of the participants as they engage with a comic page from the AI4VA dataset. 
A gaze point refers to the specific location on a visual stimulus or scene at which an individual's eyes are focused at any given moment. If a series of gaze points are close in time and spatial range, the resulting gaze cluster denotes a fixation~\cite{Konig2014Nonparametric}: a period in which our eyes are locked toward a specific region. Typically, a fixation duration is of 100-300 milliseconds. In contrast, a saccade is a rapid and involuntary eye movement that involves a quick shift in the direction of a gaze from one point to another. It is a fundamental component of normal eye movement and occurs multiple times per second during a natural visual exploration. Saccades serve the purpose of redirecting the eyes to different regions of interest in the visual field. In our work, we accumulate the gaze points to create fixations, and then aggregate all fixations from all participants for each page. With the collected data, we create the ground truth saliency maps as described in Section 3.3 in our main paper.

\subsection{Semantic Segmentation}
\begin{figure}[tb]
  \centering
  \includegraphics[width=0.48\linewidth]{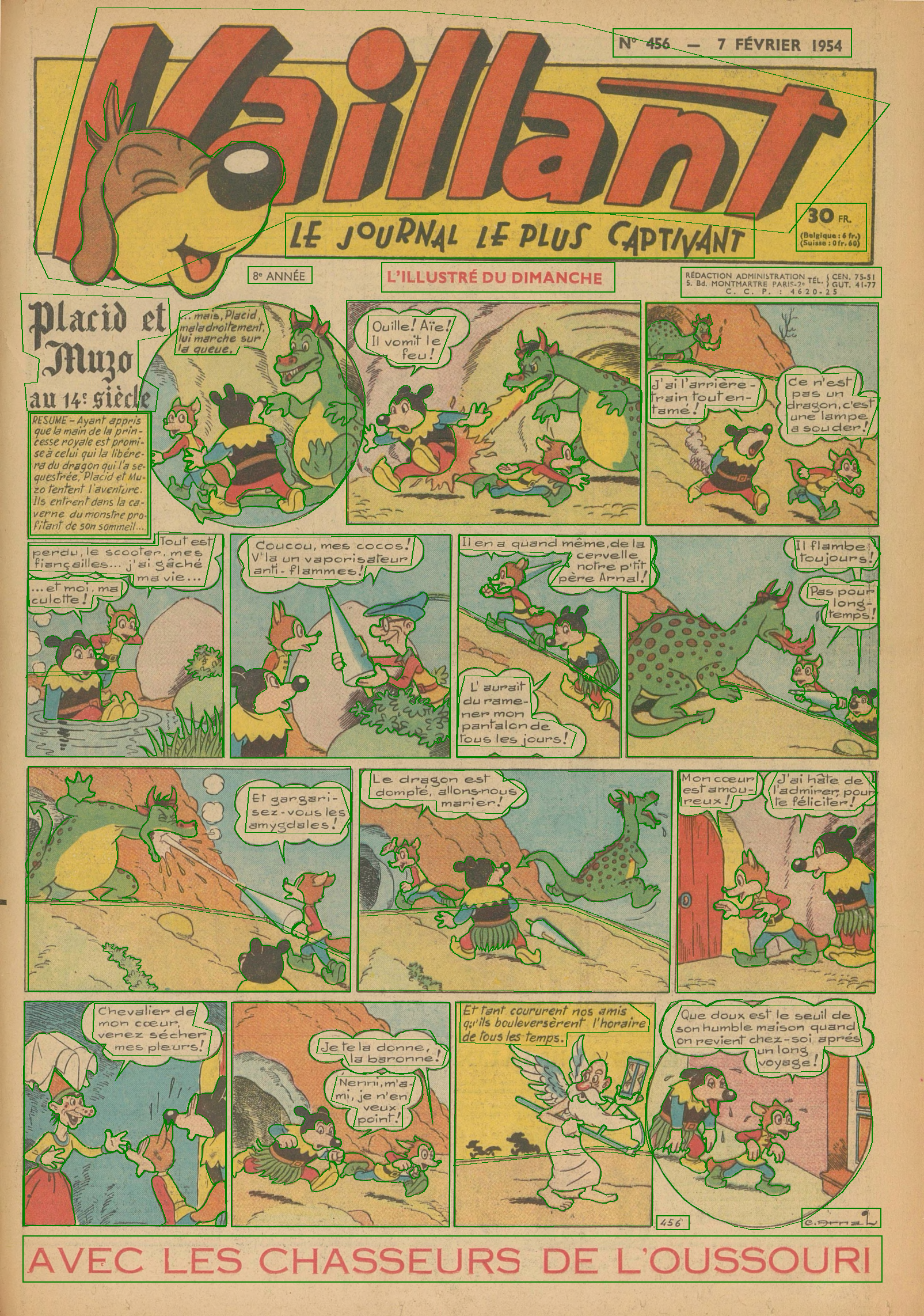}
  \includegraphics[width=0.48\linewidth]{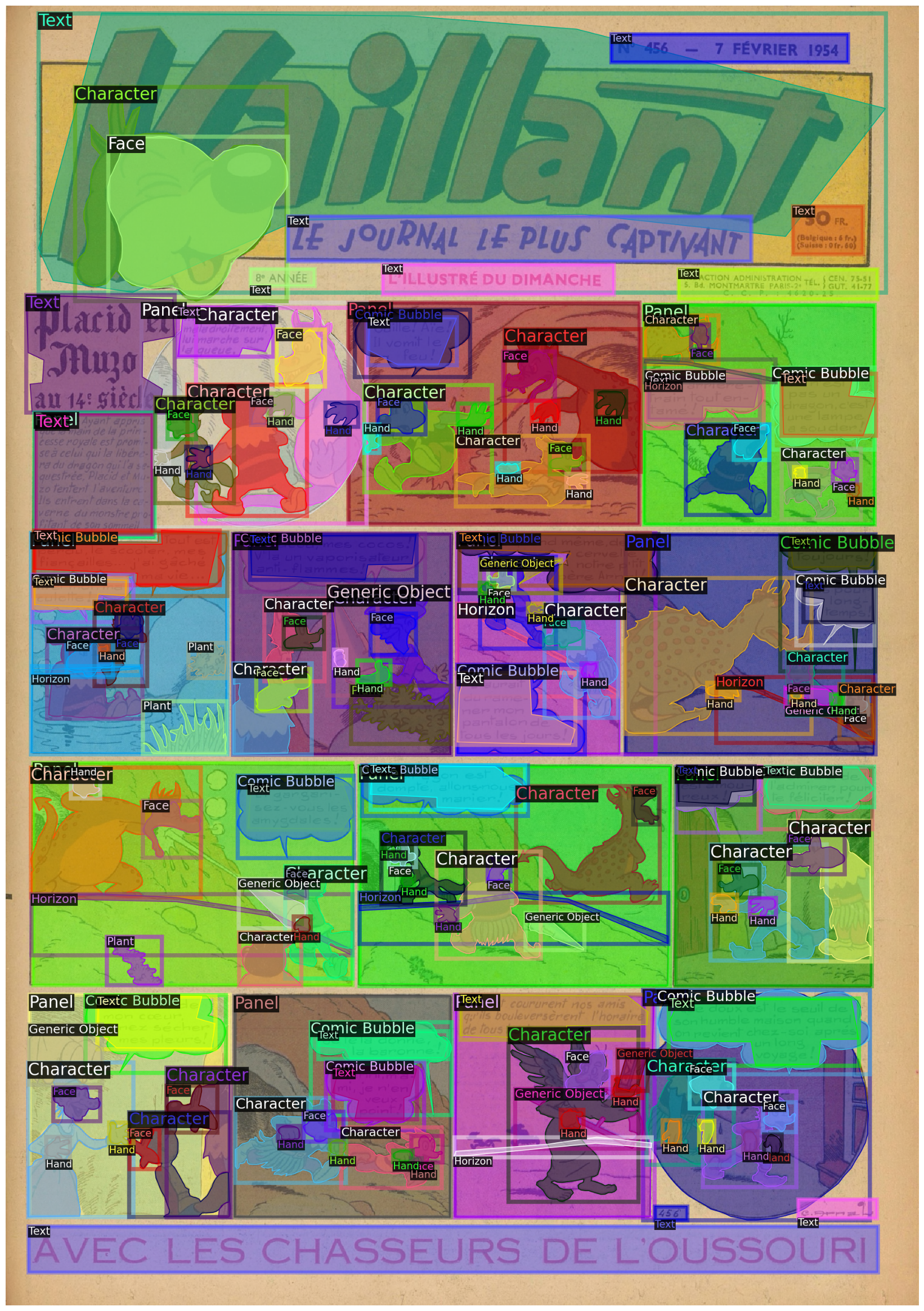}
  \caption{An annotated comic page from `Placid et Muzo' in our dataset. The dataset itself contains bounding boxes, masks and classes for instances. We note that it is compatible with Detectron2~\cite{wu2019detectron2}.}
  \label{fig:seg_page}
%  \vspace{-15pt}
\end{figure}

We provide an initial exploration into segmentation on our dataset using a pre-trained segmentation model, mask2former~\cite{cheng2022mask2former}, originally trained on the COCO dataset. Examples of ground-truth annotations from our dataset are shown in Fig.~\ref{fig:seg_page}. Note that segmentation is not included in the AI4VA workshop challenge, but future explorations in this task would be intriguing.

\section{Evaluation Metrics}

\textbf{Depth Estimation.} To evaluate the existing depth estimators for ordinal depth, we use a diverse set of metrics, described as follows. The Spearman's rank correlation coefficient ($\rho$) measures the strength and direction of the association between two ranked variables. It is suitable for ordinal data as it evaluates how well the relationship between two variables can be described using a monotonic function. 
Formally, it is represented as \[ \rho = 1 - \frac{6 \sum d_i^2}{n(n^2 - 1)}\] where $d_i$ is the difference between the ranks of corresponding values in two datasets, $n$ is the number of observations.
Further, Kendall's Tau coefficient ($\tau$) calculates the difference between the probability of concordant and discordant pairs among all possible pairs in the dataset. 
Kendall's Tau coefficient ($\tau$) is defined as \[ \tau = \frac{(\text{number of concordant pairs}) - (\text{number of discordant pairs})}{\frac{1}{2} n(n-1)} \] where concordant pairs are two pairs of observations that have the same order in both datasets and discordant pairs are two pairs of observations that have different orders in the datasets. 
Additionally, the weighted Kappa ($\kappa$) coefficient is used to measure the agreement between two raters (or annotations) on ordinal data, taking into account the order of the annotations. It assigns weights to disagreements, with more significant weights given to disagreements that are further apart. This metric is particularly useful when the ordinal categories have a natural ordering. 
Weighted Kappa ($\kappa$) is defined as \[ \kappa = 1 - \frac{\sum_{i,j} w_{ij} o_{ij}}{\sum_{i,j} w_{ij} e_{ij}} \]where $w_{ij}$ are weights assigned to the disagreement between raters for the $i$th and $j$th ratings, $o_{ij}$ is the observed frequency of ratings $i$ and $j$, $e_{ij}$ is the expected frequency of ratings $i$ and $j$, and $i$ and $j$ index the possible ratings. For all the metrics, higher values are better for performance, with +1 being the ideal.\\
\textbf{Saliency Estimation.}
To evaluate our saliency predictions, we use the metrics described as follows. Kullback-Leibler Divergence (KLD) quantifies the pixel-wise discrepancy between predicted and actual saliency distributions, with scores near zero indicating higher accuracy. Pearson’s Correlation Coefficient (CC) evaluates the linear correlation between predicted and ground-truth saliency maps, with values close to 1 denoting a strong correlation. The Similarity (SIM) score compares the minimum values across pixels between the two maps, with a score of 1 signifying a perfect match, considering both maps are normalized probability distributions.
%To evaluate saliency predictions we use the following metrics.Kullback-Leibler Divergence (KLD)~\cite{} (Vidyasagar, 2010), which encodes the cumulative pixel-wise distance between the predicted and the ground-truth saliency distributions. A KLD score close to zero indicates a better approximation of the ground-truth saliency map than the predicted one. Pearson’s correlation coefficient (CC)~\cite{} (Jost et al., 2005): This metric measures the linear relationship between the predicted and ground-truth saliency maps. It ranges from -1 to 1. A CC score close to one indicates a strong linear correlation between the two maps. \\Similarity (SIM) score~\cite{}(Judd et al., 2012): The similarity score sums, over the pixels, the minimum value between the predicted and the ground-truth saliency maps. Since both of the maps are probability distributions summing to 1, a similarity score of 1 indicates a perfect prediction.

\textbf{Semantic Segmentation.} We evaluate the existing semantic segmentation models with two evaluation methods: Page mIoU and Panel mIoU. In the former, we use the whole comic page as the input to semantic segmentation models, whereas for the latter, we run the segmentation baselines on each panel, separately. In both cases, we use the standard mIoU metric to evaluate the performance. 
\section{Baselines}
%\subsection{Semantic Segmentation}
\subsection{Depth Estimation}
We evaluate multiple depth estimators with different network architectures including convolutional neural networks, vision transformers, and diffusion models, as outlined below.
\begin{enumerate}
    \item \textbf{MIDAS}~\cite{midas}  introduces a scale and shift-invariant loss to estimate depth from a large collection of mixed real-world datasets, thereby presenting a depth model that generalizes across multiple real-world datasets.
    \item \textbf{CDE}~\cite{CDE} proposes an architecture that leverages contextual information in a given scene for monocular depth estimation. By using contextual attention, CDE obtains meaningful semantic features to enhance the performance of the depth model.
    \item \textbf{ComicDepth}~\cite{ComicDepth} is a cross-domain depth estimation method that exploits the contextual information for depth prediction of a given scene, by reasoning about global and local contexts. 
    \item \textbf{DPT}~\cite{DPT} utilizes vision transformers for dense prediction, creating multi-resolution representations from transformer stages, merged into full-resolution via a convolutional decoder. This approach, leveraging constant high-resolution and global receptive fields, yields more precise and coherent predictions than traditional fully convolutional networks.
    \item \textbf{DepthFormer}~\cite{DepthFormer} achieves monocular depth estimation by focusing on long-range correlations, combining a Transformer for global context with a convolution branch for local detail. A module for hierarchical aggregation and heterogeneous interaction links these branches, enabling direct interactions and translations between Transformer and CNN features.
    \item \textbf{ZoeDepth}~\cite{ZoeDepth} addresses single-image depth estimation, bridging the gap between relative depth estimation, which overlooks metric scale, and metric depth estimation, which often lacks generalizability. 
    \item \textbf{DDP}~\cite{DDP} presents a simple and effective framework for depth estimation through a conditional diffusion pipeline. It employs a ``noise-to-map'' approach, iteratively refining a Gaussian distribution based on the guiding image.
    \item \textbf{DepthAnything}~\cite{DepthAnything} presents a universal depth model by auto-annotating $\sim$62M unlabeled images to enhance diversity and reduce generalization errors. It uses data augmentation and auxiliary supervision to leverage semantic insights from pre-trained encoders for the task of depth estimation.
    \item \textbf{Marigold}~\cite{marigold} leverages the comprehensive priors from recent generative diffusion models, like Stable Diffusion, to enhance and generalize depth estimation. Marigold introduces an affine-invariant approach to monocular depth estimation, capitalizing on the extensive prior knowledge embedded in these models.
    \item \textbf{EVP}~\cite{EVP} utilizes the Stable Diffusion network for computer vision, featuring the Inverse Multi-Attentive Feature Refinement (IMAFR) module to improve feature learning through spatial information from upper pyramid levels. Additionally, it introduces an innovative image-text alignment module to refine feature extraction in the Stable Diffusion framework.
    
\end{enumerate}
 Note that the diffusion-based methods leverage additional training data, as outlined in their respective works~\cite{DDP, DepthAnything, marigold, EVP}.
\subsection{Saliency Estimation}
We present the results of two types of models. The first group comprises visual (human) attention methods, which aim to estimate the distribution of human attention given an input image. The second group consists of class activation mapping methods, designed to identify the regions of an image utilized by the model in its decision-making process. Despite not being trained on human attention data, several models from this second group perform competitively, as demonstrated in Table~\ref{tab:saliency-table}, suggesting a viable alternative.

Visual (human) attention methods:
\begin{enumerate}
    \item \textbf{DeepGaze I}~\cite{kummerer2015deep} uses deep learning for saliency prediction, marking a significant advancement over traditional, bottom-up models. DeepGaze I demonstrates the effectiveness of convolutional neural networks in capturing complex visual patterns relevant to human attention.
    \item \textbf{DeepGaze IIE}~\cite{linardos2021deepgaze} combines multiple image classification backbones in a principled manner and obtains good confidence calibration on unseen datasets.
    \item \textbf{DeepGaze III}~\cite{deepgaze3} combines image information with fixation history to predict where a participant might look next.
    \item \textbf{SimpleNet}~\cite{simplenet} is designed as a lightweight yet effective model for saliency prediction, focusing on achieving high performance with a simpler network architecture.
    %It emphasizes efficiency and speed, making it suitable for applications where computational resources are limited. 
    %SimpleNet demonstrates competitive accuracy in saliency prediction, proving that efficient design can coexist with high performance.
    \item \textbf{UNISAL}~\cite{unisal} is designed to generalize across different visual domains, including images, videos, and, potentially, virtual reality environments. It utilizes a unified framework, that learns domain-invariant features, enabling it to perform consistently across various types of visual content. 
    \item \textbf{TempSAL}~\cite{aydemir2023tempsal} considers the temporal nature of gaze shifts during image observation to predict saliency. It locally modulates the saliency predictions by combining the learned temporal saliency maps.
    \item \textbf{UMSI}~\cite{umsi} (Unified Model of Saliency and Importance) learns to predict visual importance in input graphic designs. It is trained on different design classes, including posters, infographics, mobile UIs, as well as natural images, and includes an automatic classification module to classify the input.  \\
\end{enumerate}
Class Activation Mapping (CAM) based methods:\\
\begin{enumerate}
    
    \item \textbf{CAM}~\cite{zhou2015cnnlocalizationcam} generates a heatmap by highlighting the areas of the image that contribute most to the model's classification decision, offering insights into the model's focus. Using CAM enables detailed saliency predictions, that are tied to the semantic understanding of the image, facilitating interpretability. We report CAM results with googlenet~\cite{szegedy2015goinggooglenet}, efficientnet\_v2\_s~\cite{efficientnet}, densenet169~\cite{DenseNet2017}, resnext101\_32x8d~\cite{xie2017aggregatedresnext}.
    \item \textbf{GradCAM}~\cite{selvaraju2017grad} highlights the gradients flowing into the final convolutional layer to understand the model's predictions. This method provides insights into the model's decision-making process, especially in the context of classification tasks. We report GradCAM results with densenet169, swin\_t~\cite{liu2021swin}, densenet161~\cite{DenseNet2017}, swin\_b~\cite{liu2021swin}.
    \item \textbf{LayerCAM}~\cite{layercam} generates saliency maps by emphasizing the contributions of individual layers to the model's final decision, offering a layer-wise perspective on saliency. This approach allows for a more granular analysis of how different levels of abstraction in the network contribute to identifying salient regions. We report LayerCAM results with densenet169~\cite{DenseNet2017}, swin\_t~\cite{liu2021swin}, swin\_b~\cite{liu2021swin}, resnext101\_32x8d~\cite{xie2017aggregatedresnext}, densenet161~\cite{DenseNet2017}.
    \item \textbf{SmoothGradCAM++}~\cite{Omeiza2019SmoothGA} refines the visualization of saliency maps by applying smoothing techniques to the gradients, resulting in more stable and interpretable visualizations. This method addresses the noise and instability often found in gradient-based visualizations. We report SmoothGradCAM++ results with densenet169 and msnanet~\cite{DenseNet2017}.
\end{enumerate}

\section{Additional Quantitative Results}
%%%segmentation quant results go here
%%%No depth results to add here.
\subsection{Semantic Segmentation}
For the task of \textbf{semantic segmentation}, we assess the performance of existing models pre-trained on the COCO dataset, by matching the COCO classes with the classes in our AI4VA dataset, based on their definitions. We use the state-of-the-art baseline of Mask2former~\cite{cheng2022mask2former} (ViT~\cite{dosovitskiy2020vit}), implemented in Detectron2~\cite{wu2019detectron2}, and report the performances on the AI4VA dataset in Table~\ref{tab:seg_performance}.

\label{sec: quantitative results}

\begin{table}[h]
\centering
\scalebox{1.0}{
\setlength{\tabcolsep}{5pt}
\begin{tabular}{lcccccc}
\textbf{Model} & \multicolumn{3}{c}{\textbf{Page mIoU}} & \multicolumn{3}{c}{\textbf{Panel mIoU}} \\ 
 & \textbf{Character} & \textbf{Text} & \textbf{Bubble} & \textbf{Character} & \textbf{Text} & \textbf{Bubble} \\ 
 \cmidrule(l){1-1}\cmidrule(l){2-4}\cmidrule(lr){5-7}
% DeepLab & 0.290 & 0.439 & 0.714 & 0.228 & 0.315 & 0.742 \\
Mask2former & 0.614 & 0.566 & 0.460 & 0.295 & 0.514 & 0.400 \\
%\hline
\end{tabular}}
\caption{Comparative analysis of semantic segmentation performance on Page mIoU and Panel mIoU metrics.}
\label{tab:seg_performance}
%\vspace{-15pt}
\end{table}
%\vspace{-15pt}
%We present the performance of semantic segmentation baselines in Table~\ref{tab:seg_performance} in terms of Page mIoU and Panel mIoU for three comic specific classes; character, text, and comic bubble. For Page mIoU, we use the whole comic page as the input to semantic segmentation models, whereas for Panel mIoU, we run the segmentation baselines on each panel, separately. ViT-based models perform better in Page mIoU, compared to Panel mIoU since global self-attention in ViT models reasons about the relationship across panels.
In Table~\ref{tab:seg_performance}, we present the performance of  mask2former~\cite{cheng2022mask2former} using Page mIoU and Panel mIoU metrics for three comic-specific classes: `character', `text', and `comic bubble'. Page mIoU evaluates whole comic pages, while Panel mIoU assesses each panel, individually. Notably, ViT-based models excel in Page mIoU, benefiting from global self-attention mechanisms that effectively capture relationships across panels. 

\subsection{Saliency Estimation}
Table~\ref{tab:saliency-table} presents additional saliency baselines not reported in the main paper. Consistent with the main results, DeepGazeIIE outperforms all other baselines across all metrics for the AI4VA dataset.
\begin{table}[h]
\centering
\begin{tabular}{lcccccc}
 & \multicolumn{3}{c}{\textbf{Page}} & \multicolumn{3}{c}{\textbf{Panel}} \\ 
\textbf{Model} & \textbf{KLD}$\downarrow$  & \textbf{CC} $\uparrow$ & \textbf{SIM}$\uparrow$  & \textbf{KLD}$\downarrow$ & \textbf{CC} $\uparrow$ & \textbf{SIM}$\uparrow$  \\ 
 \cmidrule(l){1-1}\cmidrule(l){2-4}\cmidrule(lr){5-7}
DeepGaze I & 0.290 & 0.439 & 0.714 & 0.228 & 0.315 & 0.742 \\
DeepGaze IIE & \textbf{0.207} & \textbf{0.643} & \textbf{0.771} & \textbf{0.194} & \textbf{0.615} & \textbf{0.793} \\
DeepGaze III & \underline{0.254} & 0.530 &\underline{ 0.742} & \underline{0.222} & 0.482 & 0.761 \\
SimpleNet & 0.351 & 0.548 & 0.729 & 0.783 & 0.514 & \underline{0.792} \\
UNISAL & 0.598 & 0.517 & 0.730 & 0.389 & \underline{0.528} & 0.665 \\
UMSI & 0.495 & 0.463 & 0.629 & 1.994 & 0.428 & 0.580 \\
TempSAL & 0.363 & \underline{0.573} & 0.714 & 1.329 & 0.507 & 0.555 \\

 %\cmidrule(l){1-1}\cmidrule(l){2-4}\cmidrule(lr){5-7}
CAM - googlenet & 0.373 & 0.359 & 0.696 & 0.386 & 0.312 & 0.743 \\
CAM - efficientnet\_v2\_s & 0.411 & 0.269 & 0.688 & 0.280 & 0.381 & 0.749 \\
CAM - densenet169 & 0.377 & 0.343 & 0.697 & 0.373 & 0.446 & 0.748 \\
CAM - resnext101\_32x8d & 0.534 & 0.248 & 0.674 & 0.375 & 0.430 & 0.752 \\
GradCAM - densenet169 & 0.364 & 0.351 & 0.697 & 0.375 & 0.454 & 0.751 \\
GradCAM - swin\_t & 0.432 & 0.395 & 0.700 & 0.711 & 0.318 & 0.711 \\
GradCAM - densenet161 & 0.453 & 0.327 & 0.689 & 0.357 & 0.441 & 0.751 \\
GradCAM - swin\_b & 0.695 & 0.374 & 0.675 & 0.798 & 0.477 & 0.732 \\
LayerCAM - densenet169 & 0.350 & 0.370 & 0.703 & 0.338 & 0.452 & 0.755 \\
LayerCAM - swin\_t & 0.414 & 0.387 & 0.701 & 0.604 & 0.300 & 0.713 \\
LayerCAM - swin\_b & 0.684 & 0.374 & 0.679 & 0.743 & 0.379 & 0.738 \\
LayerCAM - resnext101\_32x8d & 0.481 & 0.270 & 0.682 & 0.368 & 0.432 & 0.760 \\
LayerCAM - densenet161 & 0.392 & 0.337 & 0.693 & 0.325 & 0.447 & 0.758 \\
SmoothGradCAM++ - densenet169 & 0.333 & 0.398 & 0.713 & 0.330 & 0.464 & 0.760 \\
SmoothGradCAM++ - msnanet1.0 & 0.392 & 0.336 & 0.698 & 0.316 & 0.496 & 0.765 \\
\end{tabular}
\caption{Comparative analysis of model performance on page and panel-based evaluations.   The results in bold and italics show the best and the second-best performance respectively. DeepGaze IIE~\cite{linardos2021deepgaze} outperforms other models in both page-wise and panel-wise evaluations.}
\label{tab:saliency-table}
\end{table}

\section{Additional Qualitative Results}
\subsection{Semantic Segmentation}

\begin{figure}[h]
  \centering
  \includegraphics[width=\linewidth]{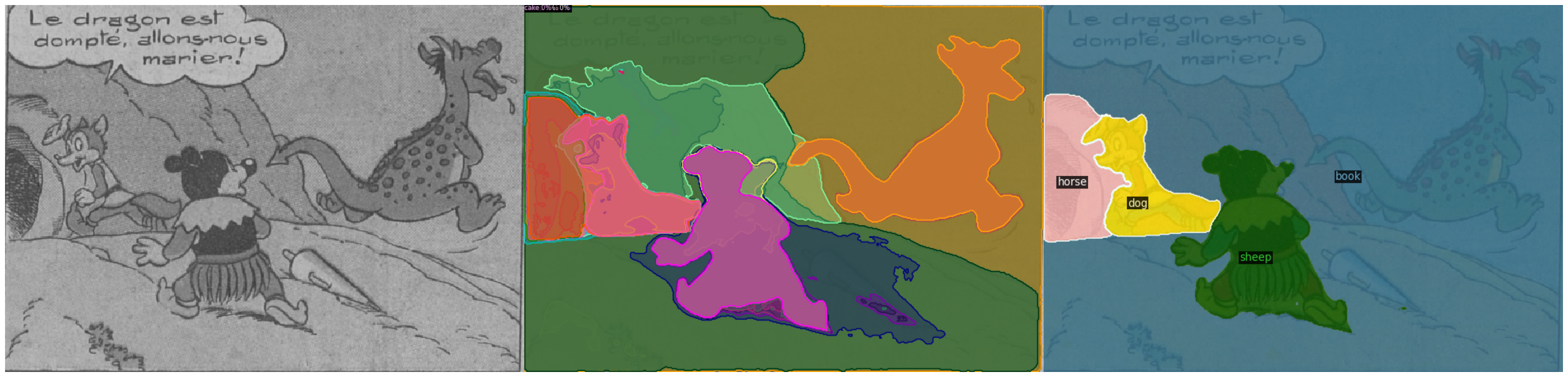}
  \includegraphics[width=\linewidth]{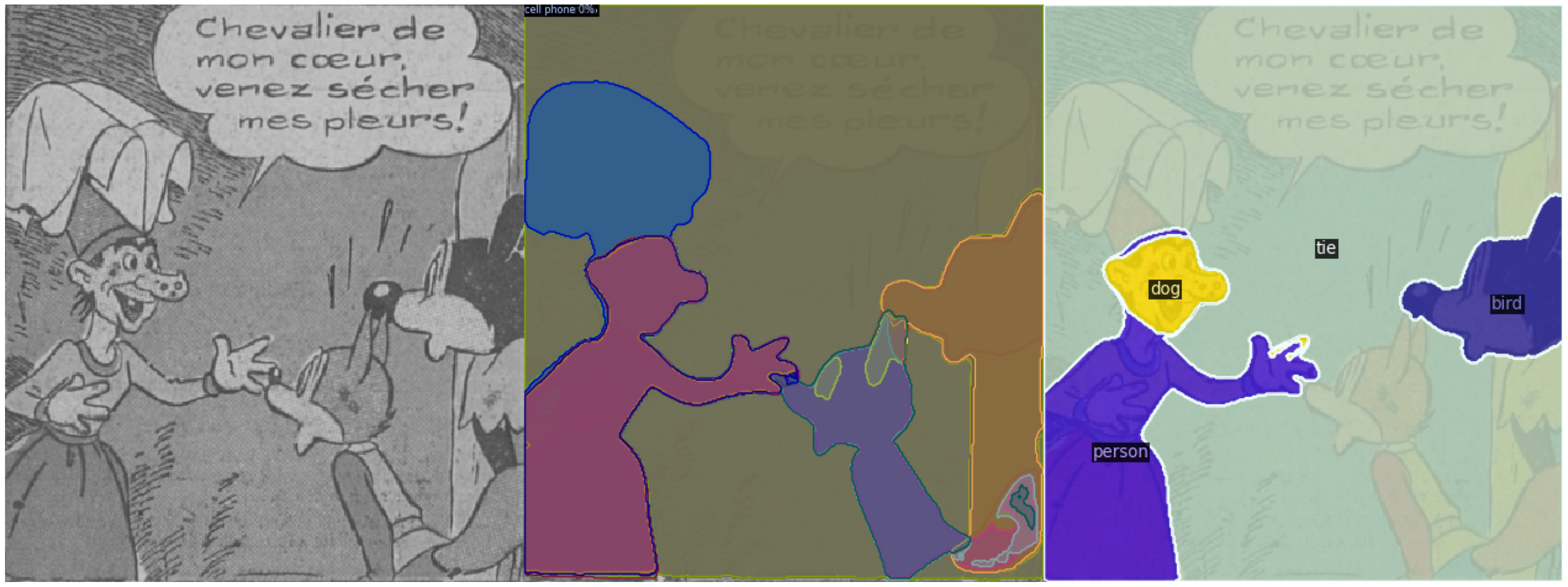}
  \\
  \hspace{40pt} Panel \hspace{55pt} Instance Segmentation \hspace{15pt} Semantic Segmentation
 \caption{Segmentation predictions by mask2former~\cite{cheng2022mask2former} on a panel from "Placid et Muzo". While the model segments objects and classes relatively well, the class assignments are understandably incorrect and imprecise. Therefore, we do not enforce segmentation tasks using fine-tuned or supervised models within the AI4VA challenge. Instead, we present this work to lay the foundation for semantic understanding in comics, with the potential for future research using language-based semantic models. On the right, the characters are misclassified into different classes such as person, bird, sheep, and dog.}
  \label{fig:seg_qualt}
%  \vspace{-15pt}
\end{figure}

Qualitative results of the COCO pretrained models on our dataset in Fig~\ref{fig:seg_qualt} demonstrate that these models can segment the objects but fail to classify them. In two examples shown in Fig.~\ref{fig:seg_qualt}, we can see that characters are segmented accurately but misclassified with various classes; bird, person, dog, and sheep.

\subsection{Depth Estimation} 
In Fig 3, we show that for depth estimation, the models predict the orders in alignment with that of the ground-truth depth annotations, wherein DepthAnything~\cite{DepthAnything} outperforms the ZoeDepth~\cite{ZoeDepth} method. As DepthAnything is a diffusion-based method it leverages extra text-image cues to predict better ordinal depth as compared to the transformer-based ZoeDepth. Notably, the DepthAnything model can predict the ordinal depth of fine structures like the hand and the snake in the bottom rows of the `Placid et Muzo' results, respectively. Nonetheless, the ZoeDepth baseline model can predict the ordinal depth of larger graphical structures, that align with the ground-truth depth annotations of AI4VA. We present the qualitative results of the two best-performing models across the employed methods of Diffusion~\cite{DepthAnything} and Vision Transformers~\cite{ZoeDepth}, respectively.
 \begin{figure}[tb]
  \centering
  \includegraphics[width=\linewidth]{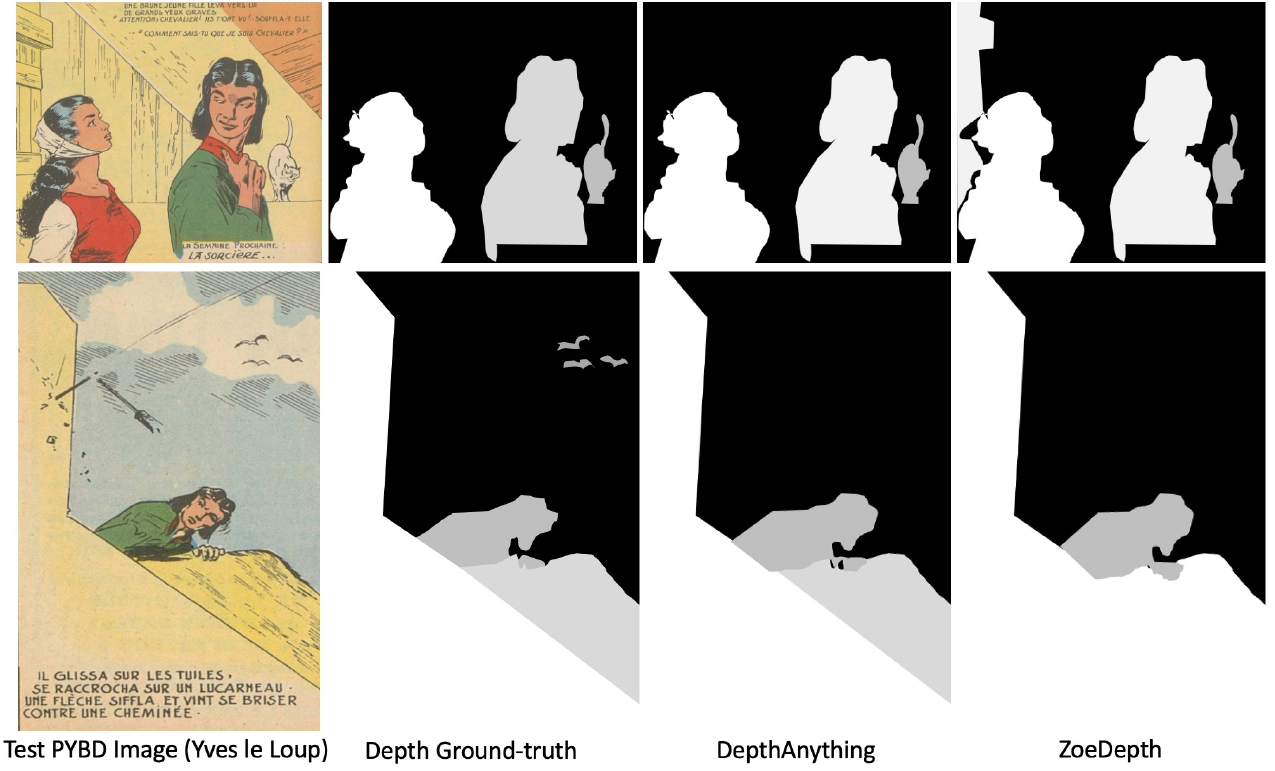}
  %%Placid results go here
  \includegraphics[width=\linewidth]{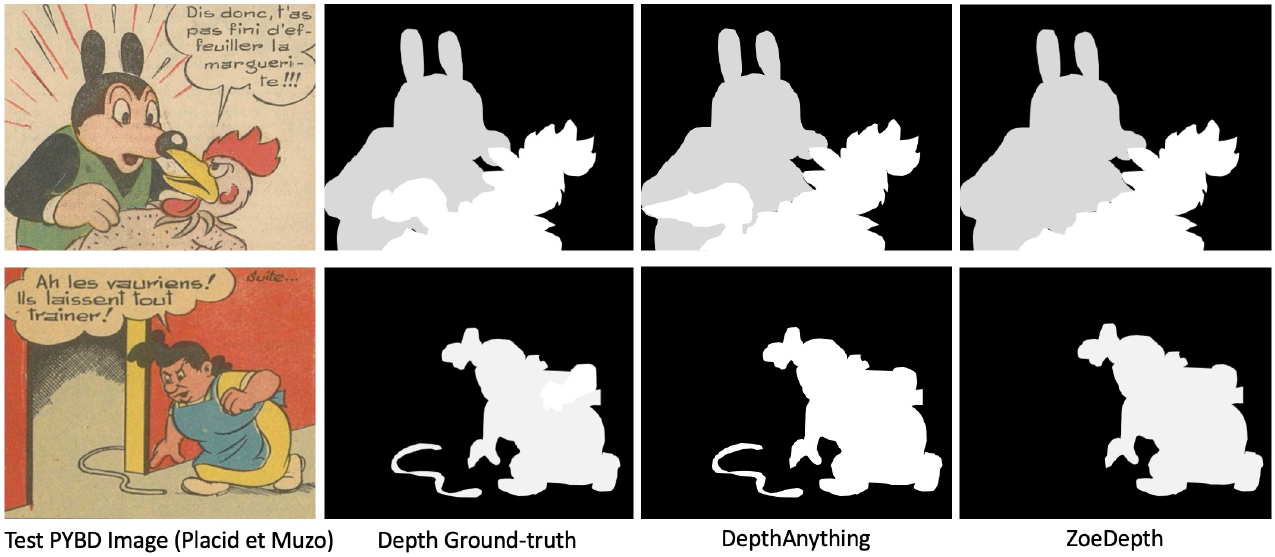}
  \caption{Qualitative results for depth ordering on AI4VA test images from `Yves le Loup' (top two rows) and `Placid et Muzo' (bottom two rows), respectively. Our depth annotations align well with the predictions from ordinal depth methods like DepthAnything~\cite{DepthAnything} and ZoeDepth~\cite{ZoeDepth}.}
  \label{fig:qualitative depth result}
  \vspace{-15pt}
\end{figure}

\subsection{Saliency Estimation}
In Fig.4 and 5, we present the saliency predictions from DeepGaze IIE\cite{linardos2021deepgaze} and SimpleNet~\cite{simplenet} methods along our ground truth data. DeepGaze IIE demonstrates a superior ability to identify the saliency of overall panel structure and textual elements. Conversely, SimpleNet is better at recognizing faces as salient regions, as evidenced in Fig. 4(d) and Fig. 5d). However, it's important to note that saliency is inherently context-dependent; a single panel can overshadow others, as seen in Fig. 4(f). Interestingly, participants did not allocate substantial attention to the character 'Placid' in Fig.5(e), likely because the text directs attention towards the footprints on the floor. This results in an overestimation of the saliency of this character by both models as seen in Fig.5(g) and Fig.5(h).

\begin{figure}
        \captionsetup{justification=centering}  % Center the caption

    \centering
        \begin{subfigure}{0.23\textwidth}
        \centering
        \includegraphics[width=\linewidth]{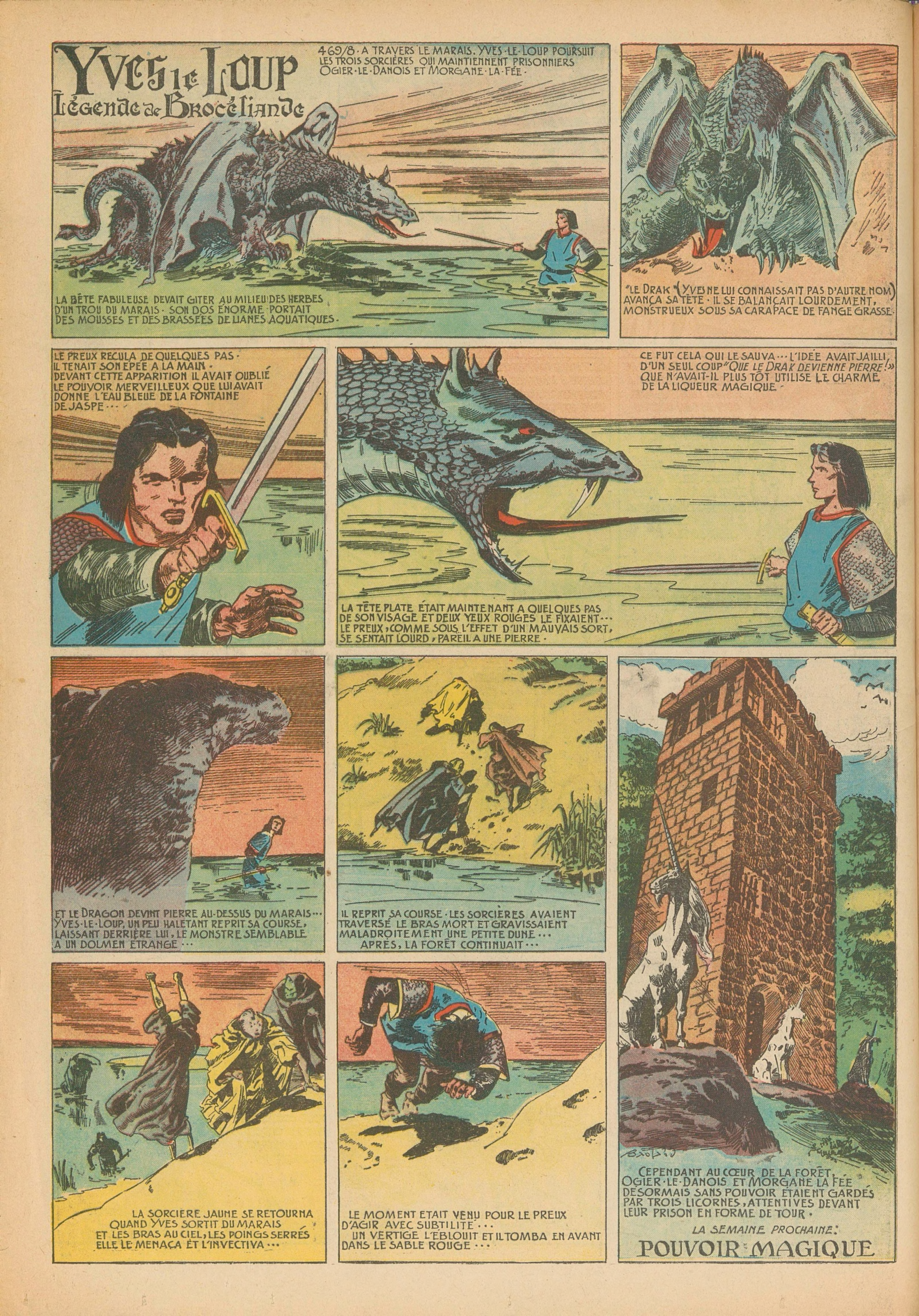}
        \caption{Test PYPD Page (Yves de Loup)}
    \end{subfigure}
    \begin{subfigure}{0.23\textwidth}
        \centering
        \includegraphics[width=\linewidth]{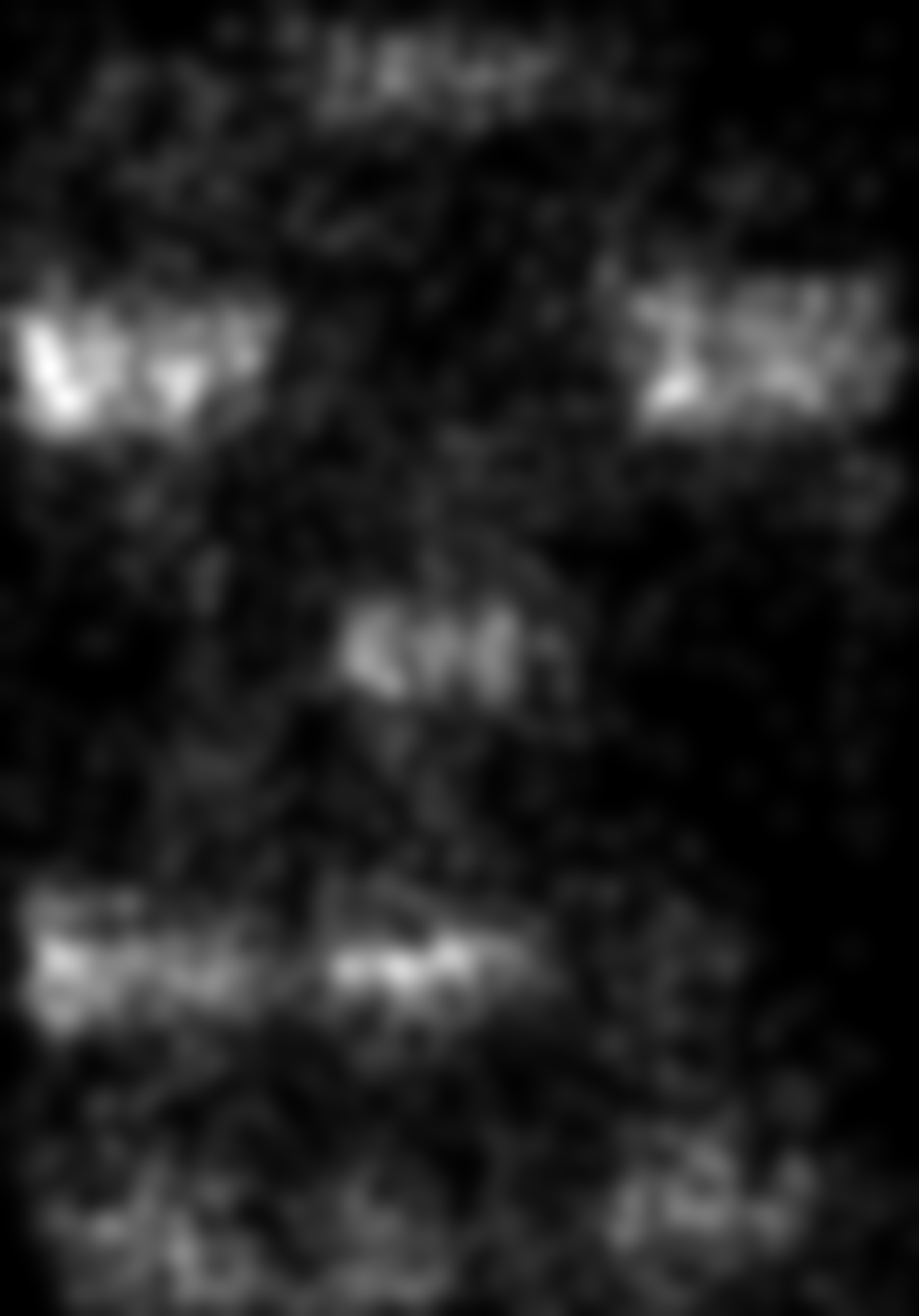}
        \caption{Saliency Ground Truth}
    \end{subfigure}
    \begin{subfigure}{0.23\textwidth}
        \centering
        \includegraphics[width=\linewidth,height=4.01cm]{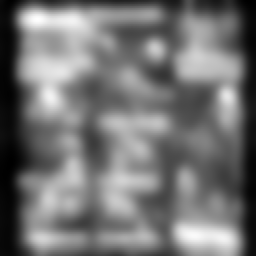}
        \caption{DeepGazeIIE Prediction}
    \end{subfigure}
    \begin{subfigure}{0.23\textwidth}
        \centering
        \includegraphics[width=\linewidth,height=4.01cm]{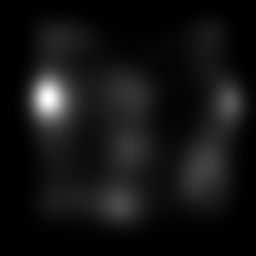}
        \caption{Simplenet Prediction}
    \end{subfigure}
    \begin{subfigure}{0.23\textwidth}
        \centering
        \includegraphics[width=\linewidth]{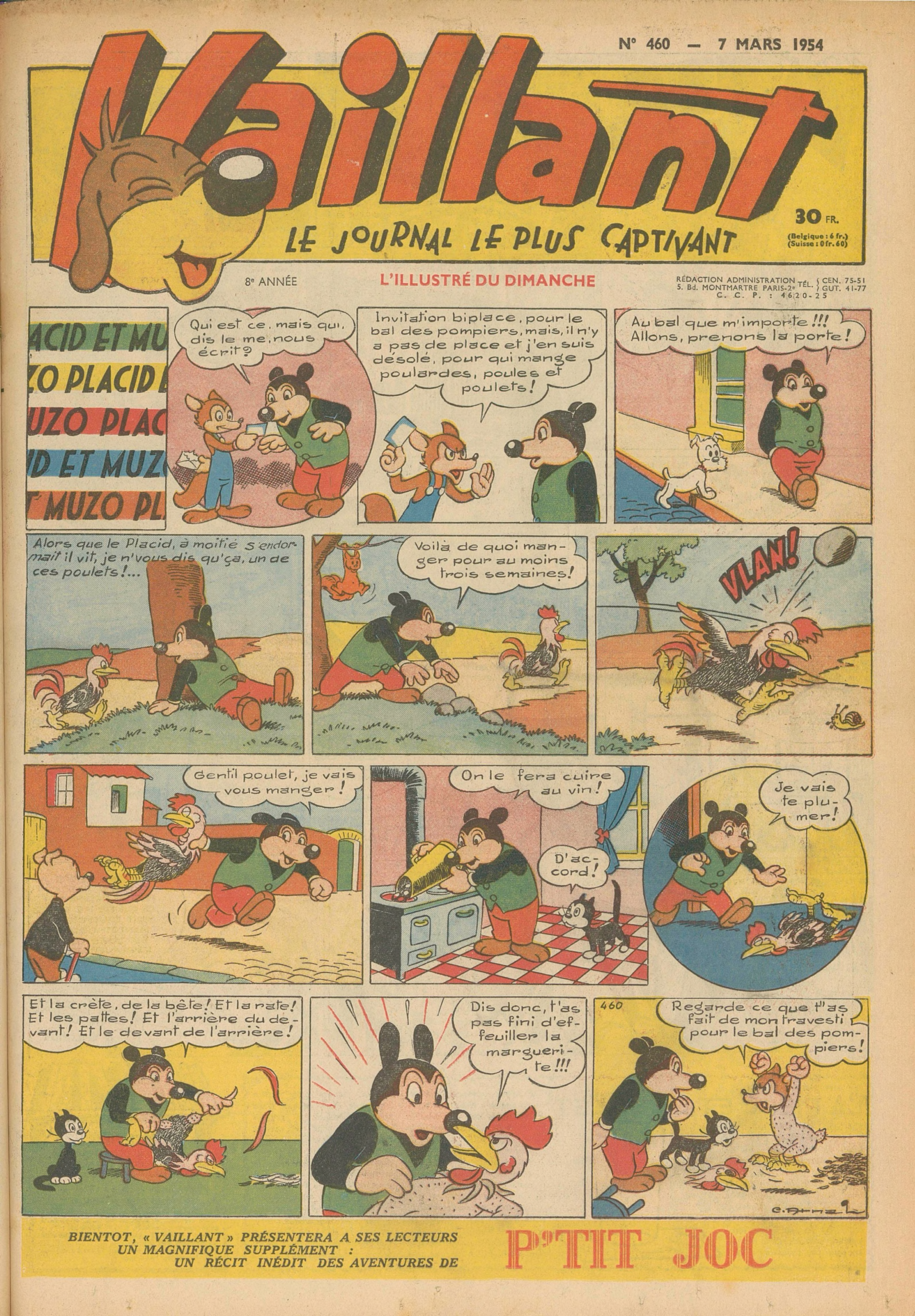}

        \caption{Test PYPD Page (Placid et Muzo)}
    \end{subfigure}
    \begin{subfigure}{0.23\textwidth}
        \centering
        \includegraphics[width=\linewidth]{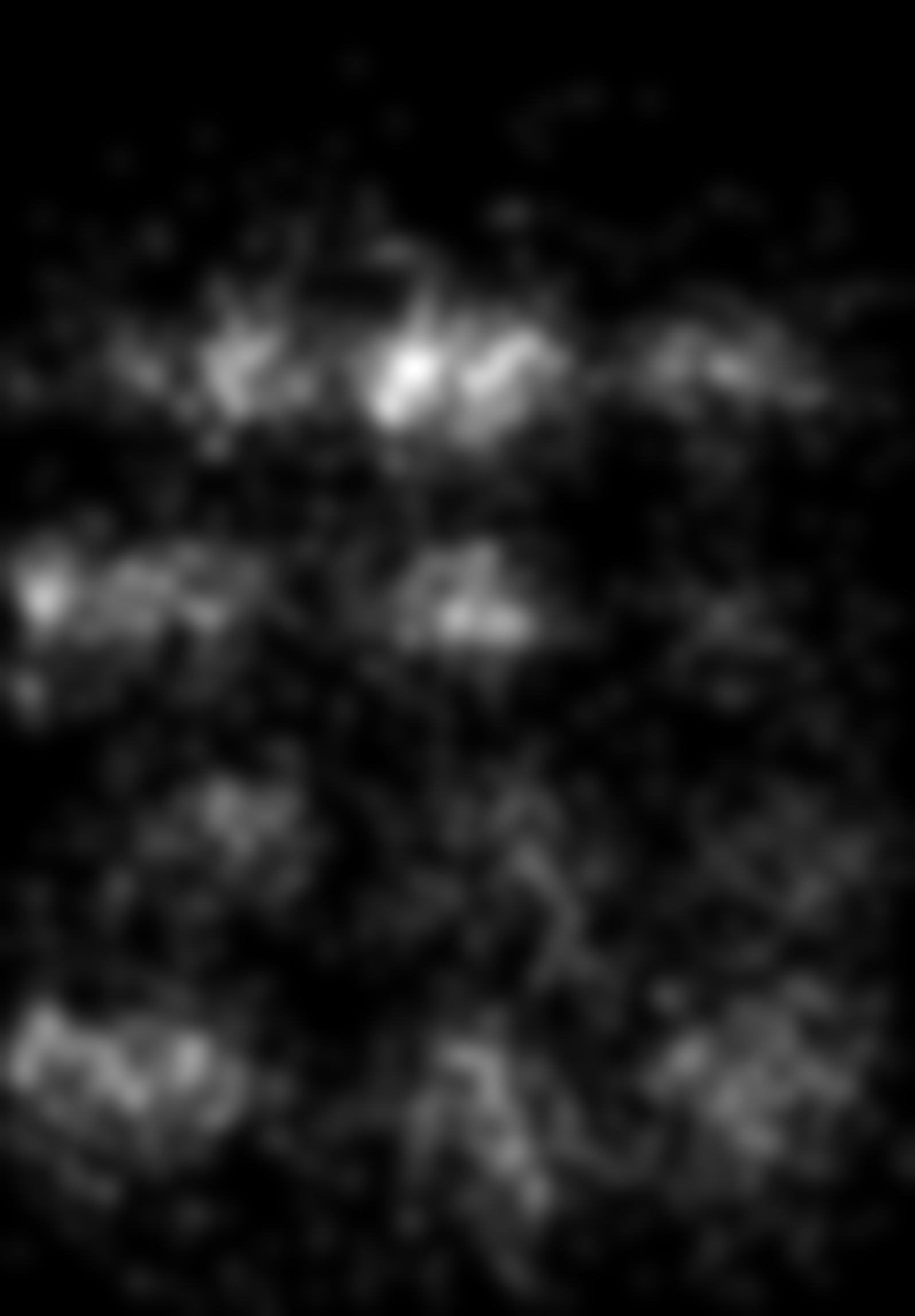}
        \caption{Saliency Ground Truth}
    \end{subfigure}
    \begin{subfigure}{0.23\textwidth}
        \centering
        \includegraphics[width=\linewidth,height=4.01cm]{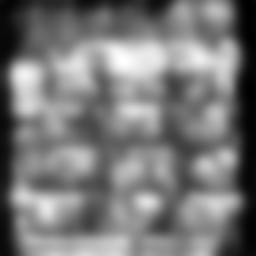}
        
        \caption{DeepGazeIIE Prediction}
    \end{subfigure}
    \begin{subfigure}{0.23\textwidth}
        \centering
        \includegraphics[width=\linewidth,height=4.01cm]{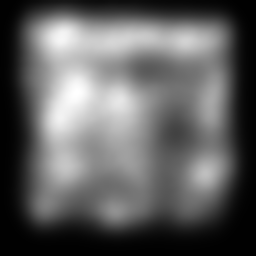}
        \caption{Simplenet Prediction}
    \end{subfigure}
    \\

            \captionsetup{justification=justified}  % Center the caption

    \caption{Qualitative results for saliency on AI4VA test pages from `Yves le Loup' (top row) and `Placid et Muzo' (bottom row), respectively. }
   % \vspace{-25pt}
\end{figure}
%\vspace{-20pt}
\begin{figure}
        \captionsetup{justification=centering}  % Center the caption

    \centering
    \begin{subfigure}{0.23\textwidth}
        \centering
        \includegraphics[width=\linewidth]{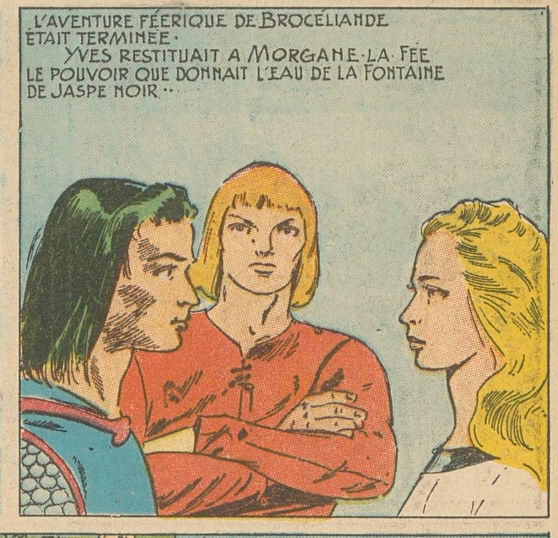}
        \caption{Test PYPD Panel (Yves de Loup)}
    \end{subfigure}
    \begin{subfigure}{0.23\textwidth}
        \centering
        \includegraphics[width=\linewidth]{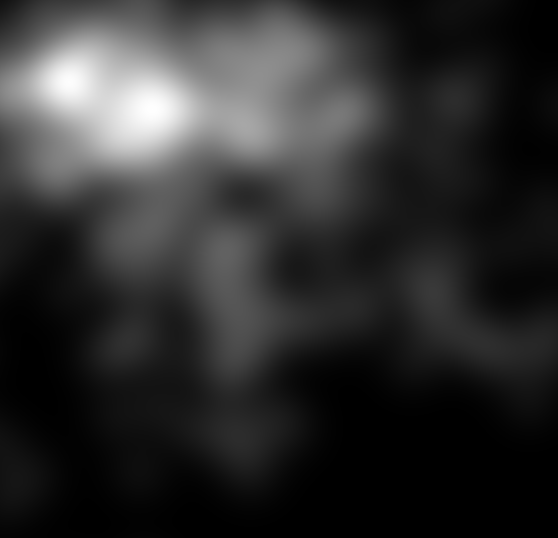}
        \caption{Saliency Ground Truth}
    \end{subfigure}
    \begin{subfigure}{0.222\textwidth}
        \centering
        \includegraphics[width=\linewidth]{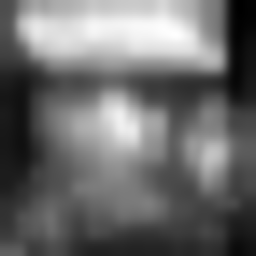}
        \caption{DeepGazeIIE Prediction}
    \end{subfigure}
    \begin{subfigure}{0.23\textwidth}
        \centering
        \includegraphics[width=\linewidth,height=0.97\linewidth]{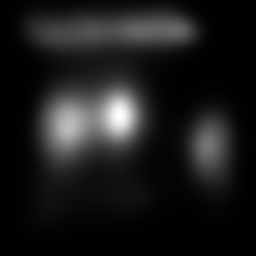}
        \caption{Simplenet Prediction}
    \end{subfigure}
    \\
    \begin{subfigure}{0.23\textwidth}
        \centering
        \includegraphics[width=\linewidth]{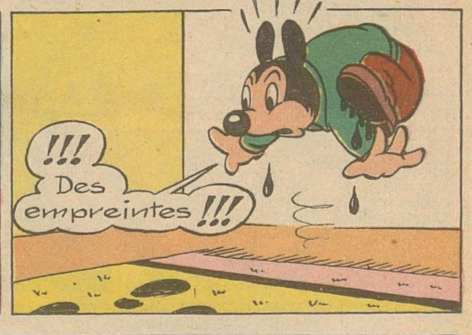}
        \caption{Test PYPD Panel (Placid et Muzo)}
    \end{subfigure}
    \begin{subfigure}{0.23\textwidth}
        \centering
        \includegraphics[width=\linewidth]{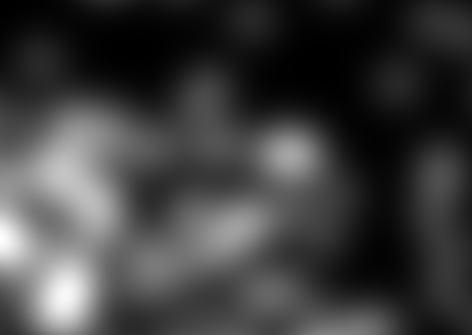}
        \caption{Saliency Ground Truth}
    \end{subfigure}
    \begin{subfigure}{0.23\textwidth}
        \centering
        \includegraphics[width=\linewidth,height=0.71\linewidth]{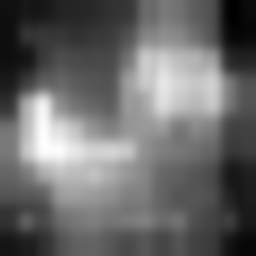}
        \caption{DeepGazeIIE Prediction}
    \end{subfigure}
    \begin{subfigure}{0.23\textwidth}
        \centering
        \includegraphics[width=\linewidth,height=0.71\linewidth]{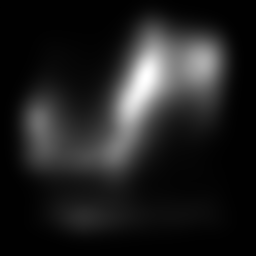}
        \caption{Simplenet Prediction}
    \end{subfigure}
            \captionsetup{justification=justified}  % Center the caption

    \caption{Qualitative results for saliency on AI4VA test panels from `Yves le Loup' (top row) and `Placid et Muzo' (bottom row), respectively. Although the models exhibit limited performance, our saliency ground truth aligns well with the textual content, panels, and objects. }
\end{figure}\label{fig:saliency-image}
\clearpage  % TODO REVIEW/FINAL: This \clearpage needs to be removed from both review and camera-ready versions.

%\bibliographystyle{splncs04}
%\bibliography{egbib}